\pdfoutput=1

\documentclass[11pt]{article}

\usepackage[]{ACL2023}

\usepackage{times}
\usepackage{latexsym}

\usepackage[T1]{fontenc}

\usepackage[utf8]{inputenc}

\usepackage{microtype}

\usepackage{inconsolata}
\usepackage{multirow}
\usepackage{graphicx}
\usepackage{float}
\usepackage{soul}
\usepackage{algorithm}
\usepackage{algpseudocode}
\usepackage{amsthm,amsmath,amsfonts,bm,xspace}
\usepackage{color}

\def\eg{{\em e.g.,}\xspace}
\def\ie{{\em i.e.,}\xspace}

\newcommand{\comment}[1]{}

\definecolor{mgreen}{rgb}{0,0.7,0}

\def\eqref#1{(\ref{#1})}

\def\1{\bm{1}}

\def\vc{{\bm{c}}}

\def\vz{{\bm{z}}}

\def\mC{{\bm{C}}}

\DeclareMathAlphabet{\mathsfit}{\encodingdefault}{\sfdefault}{m}{sl}
\SetMathAlphabet{\mathsfit}{bold}{\encodingdefault}{\sfdefault}{bx}{n}

\def\gE{{\mathcal{E}}}

\def\gG{{\mathcal{G}}}
\def\gH{{\mathcal{H}}}

\def\gS{{\mathcal{S}}}

\def\gU{{\mathcal{U}}}
\def\gV{{\mathcal{V}}}

\def\sD{{\mathbb{D}}}

\newcommand{\indep}{\perp \!\!\! \perp}
\newcommand{\dep}{\not\!\perp\!\!\!\perp}

\newcommand{\system}[1]{\text{#1}}

\newcommand{\roberta}{\system{RoBERTa}\xspace}
\newcommand{\blenderbot}{\system{Blenderbot}\xspace}
\newcommand{\constrain}{\system{ConSTrain}\xspace}
\newcommand{\alpaca}{\system{Alpaca}\xspace}
\newcommand{\ourmethod}{\system{CausalScore}\xspace}

\title{\ourmethod: An Automatic Reference-Free Metric for Assessing Response Relevance in Open-Domain Dialogue Systems}

\author{Tao Feng \and Lizhen Qu \and  Xiaoxi Kang \and Gholamreza Haffari \\
        Monash University, Australia \\ \texttt{firstname.lastname@monash.edu} \\ }

\begin{document}
\maketitle
\begin{abstract}
Automatically evaluating the quality of responses in open-domain dialogue systems is a challenging but crucial task. Current evaluation metrics often fail to align with human judgments, especially when assessing responses that are grammatically correct. To address this issue, we propose a novel metric, called \ourmethod, which assesses the relevance of responses by measuring the causal strength between dialogue histories and responses. The causal strength is estimated by utilizing both unconditional dependence and conditional dependencies from the dialogue history to responses. We compare our metric with the existing competitive metrics in terms of their alignment with human judgements. Our experimental results demonstrate that \ourmethod significantly surpasses existing state-of-the-art metrics by aligning better with human judgements. Additionally, we collect a new dialogue dataset CGDIALOG+ with human-annotated causal relations and a set of pairwise human judgements to facilitate the development of future automatic metrics. \footnote{Codes and datasets are available at \url{https://github.com/WilliamsToTo/causalscore_dialogue}.} 
\end{abstract}

\section{Introduction}
Although various automatic metrics~\cite{papineni-etal-2002-bleu,lin2004rouge, tao2018ruber, ghazarian-etal-2022-deam} have been proposed in the past, evaluation of open-domain dialogue systems is still an open challenge. Existing metrics often show a low correlation with human judgements~\cite{ma2023ffaeval}. In particular, assessing to what degree a response is semantically relevant to the corresponding dialogue history is a difficult task.

Reference-based metrics, such as BLEU~\cite{papineni-etal-2002-bleu} and BERTScore~\cite{BERTScore}, assess the quality of generated dialogue responses by measuring their similarities to human written ``gold'' responses. However, they cannot accurately and impartially evaluate \textit{diverse} texts generated by the systems built upon large language models (LLMs), especially when the responses differ significantly from references but are still plausible and fluent for humans~\cite{liu2023gEval}. 

\begin{figure}
    \centering
    \includegraphics[width=\linewidth]{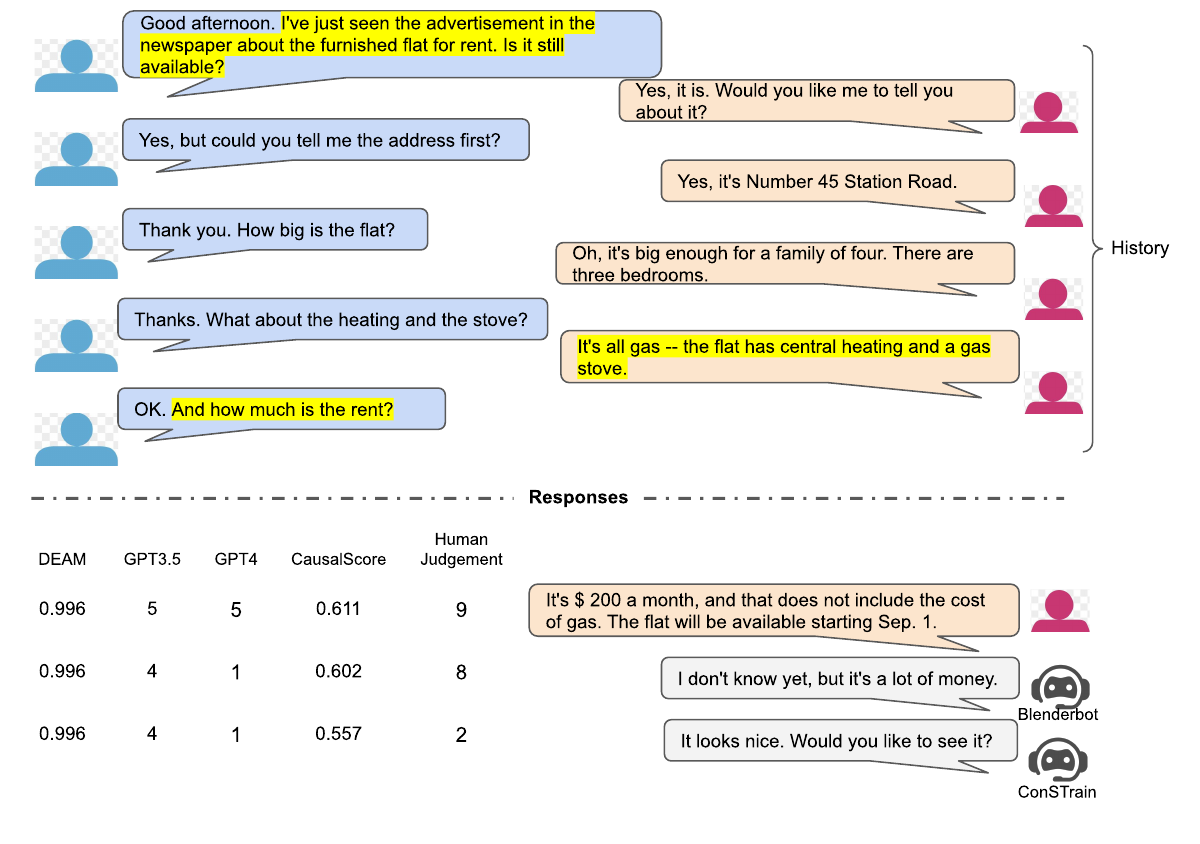}
    \caption{This is an illustrative example of dialogue evaluation, where the responses are generated by human and different dialogue systems. Evaluation results for \textbf{relevance} using different metrics are provided alongside the responses. \hl{Highlighted texts} indicate causes of human response.}
    \label{fig:causal_relation_example}
\vspace{-4mm}
\end{figure}

In contrast, reference-free metrics are proposed to directly output scores based on the dialogue history and responses without the references. There are in general two paradigms to evaluate responses from dialogue models: i) supervised models, which are classifiers or regression models to estimate a score for a given response, such as ADEM~\cite{lowe-etal-2017-towards}, RUBER~\cite{tao2018ruber}, and DEAM~\cite{ghazarian-etal-2022-deam}, and ii) pre-trained LLMs, which are employed to generate a score indicating the quality of a response~\cite{liu2023gEval}. However, as illustrated in Fig. \ref{fig:causal_relation_example}, our study (see Sec. \ref{sec:results}) reveals that these metrics frequently assign high scores to grammatically correct responses, but none of those scores correlate well with the corresponding human rankings on crucial evaluation aspects (\eg relevance, empathy, etc) even in the in-domain setting.  

Based on the above analysis, this work focuses on developing an automatic, reference-free metric that better aligns with human judgements in evaluating the relevance of responses. \citet{feng-etal-2023-less} show that responses which are highly relevant to the dialogue history also exhibit a strong causal relation between the history and the responses. As shown in Fig~\ref{fig:causal_relation_example}, the most relevant response (\ie the human response) replies to more utterances in the dialogue history. For instance, the question "how much is the rent?" causes the response containing "It's \$200 a month". Similarly, because the history states "It's all gas - the flat has central heating and a gas stove," the human responds with "That does not include the cost of gas." Additionally, the question "Is it still available?" elicits the response "The flat will be available starting September 1." However, other responses have few or no such causal relations. Inspired by this finding, we propose a novel metric \ourmethod to quantify the relevance of responses by estimating the causal strength~\cite{janzing2013quantifying} between utterances and responses, where causal strength measures the strength of causal relations. Namely, a response assigned with a high causal strength score indicates it is highly relevant to dialogue history. 

We use classifier-based (un)conditional independence tests to estimate causal strength~\cite{bookCausationSpirtes, 10.5555/1642718, mukherjee2020ccmi}. Specifically, the implementation of \ourmethod involves a three-step process. First, we apply an unconditional independence classifier to identify a subset of the utterances in dialogue history that statistically depend on a given response, named dependent utterances. Second, we calculate conditional dependencies using the conditional independence classifier, which is operated by conditioning each utterance in dependent utterances. Finally, \ourmethod estimates causal strength by aggregating both unconditional and conditional dependencies. In contrast, \citet{feng-etal-2023-less} only computes the highest conditional dependence conditioned on the preceding utterance of responses.

To train classifiers in \ourmethod, we extend the CGDIALOG dataset~\cite{feng-etal-2023-less} with a new domain DREAM~\cite{gu-etal-2022-dream} to build a new corpus, coined CGDIALOG +, which contains a total of 2,444 history-response pairs across three domains. Each dialogue is annotated with causal relations between history and response. To evaluate the alignment between an automatic metric and human judgements, we ask crowd-workers to indicate their preference between a pair of responses, given the same dialogue history. This ends up with 1,800 annotated preferences, which are used to conduct extensive experiments to compare \ourmethod with the state-of-the-art (SOTA) automatic metrics. The experimental results show that \ourmethod has significantly stronger correlations with human judgements than the SOTA metrics by using a variety of correlation measures.



\section{Background}
\label{sec:background}

\paragraph{Causal Discovery and Causal Strength.}
Unlike traditional statistical analysis, which focuses on correlation analysis between variables, causal discovery aims to discover a causal graph among a set of variables through data. A causal graph $\gG$ consists of a set of nodes $\gV$ and a set of edges $\gE$, where a node $v \in \gV$ denotes a random variable and a directed edge $v_i \rightarrow v_j \in \gE$ indicates that $v_i$ is a \textit{direct cause} of $v_j$~\cite{pearl2009causal, neal2020causalitybook}. Causal discovery algorithms can be roughly divided into two categories: constraint-based method and score-based method~\cite{bookCausationSpirtes, 10.5555/3087158.3087202, pearl2009causal}.  One widely-used constraint-based causal discovery algorithm is the Peter-Clark (PC) algorithm~\cite{bookCausationSpirtes}.

For a pair of variables $(v_i, v_j)$, the PC algorithm operates unconditional independence tests and conditional independence (CI) tests given the other variables. If $v_i$ and $v_j$ are independent according to any of the tests, the PC algorithm concludes that there is no causal relation between $v_i$ and $v_j$.
The orientation of edges is determined using heuristics and identifying the specific structure such as immorality~\cite{pearl2009causal, neal2020causalitybook}.


The core of the PC algorithm is the CI test. Given $\mathit{n}$ i.i.d samples from the distribution $P(v_i,v_j,v_k)$, we say that $v_i$ is conditionally independent of $v_j$ given $v_k$ (denoted by $v_i\indep v_j|v_k$), if the distribution $P(v_i,v_j|v_k)$ factories as $P(v_i|v_k)P(v_j|v_k)$. 
The resulting hypothesis testing is as follows: 
Given $\mathit{n}$ i.i.d samples from the distribution $P(v_i,v_j,v_k)$, one needs to distinguish between the two hypotheses: $$\gH_{0}: v_i\indep v_j|v_k \ \ \ \textrm{vs} \ \ \  \gH_{1}: v_i\dep v_j|v_k.$$ 
Conditional independence tests can also be operationalised or interpreted based on conditional mutual information (CMI) \cite{10.5555/1146355, 138734f0-eb1f-3daf-a1d4-a16462fb1be7, pmlr-v115-mukherjee20a}, because CMI is zero if two variables are conditional independent, otherwise CMI is proportional to the dependencies between two variables. Thus, prior works also use CMI as an indicator of causal strength~\cite{seitzer2021causal}.


\paragraph{Classifier-based CI Test.} 
%

There are many CI tests for statistical data, such as Fisher-z test ~\cite{Fisher014OT}, Chi-Square test~\cite{McHugh2013-dp}, and kernel-based CI test~\cite{10.5555/3020548.3020641}. However, those methods are designed for continuous random variables, and cannot be directly applied to text data. Classifier-based CI tests convert the CI test into a binary classification problem~\cite{lopez-paz2017revisiting, 10.5555/3294996.3295055,sen2018mimic,pmlr-v115-mukherjee20a}. The central idea is to train a binary classification model to identify whether data examples are from $v_i \indep v_j|v_k$ or $v_i \dep v_j|v_k$. In this work, we adopt classifier-based CI tests to text data to identify causal relations and compute causal strength between dialogue history and response.

\section{Methodology}
\label{sec:methodology}
In this paper, we propose a reference-free automatic evaluation metric, named \ourmethod, to assess the relevance of a given response to the corresponding dialogue history. 
Formally, we are given a dialogue history $\vc=\{ c_1, ..., c_{t-1} \}$ and a response $r_t$, where each $c_i$ is an utterance in the history. The goal is to develop a function $f: (\vc, r_t) \rightarrow s$ that produces a score $s$ indicating their causal strength. We argue that \emph{a response exhibiting high relevance to the dialogue history inherently entails strong causal strength with a particular set of utterances in that dialogue history.} 

To quantify causal strength between utterances and responses, we integrate the classifier-based (un)conditional test results into a single score, inspired by the PC algorithm. By using a procedure similar to the PC algorithm, the more causal relations we find between a response and utterances, the stronger the causal strength is. We run first unconditional tests to identify strong candidates of causal relations, followed by verifying them with CI tests. Both types of tests are conducted by employing a classifier, which predict the probability of being dependent between a response and input utterances. Instead of discovering full causal graphs, we average among these dependence classifier probabilities based on the selected candidates after unconditional tests to produce the final score.

In the following, we first introduce the CGDIALOG+ corpus, followed by how we build the classifiers on that corpus and employ their predictions to calculate \ourmethod. 
%
                                                                %



\subsection{CGDIALOG+}
CGDIALOG+ is an extension of CGDIALOG that is used to train the classifiers for independence tests. CGDIALOG is a dialogue dataset with human-annotated causal relations between utterances in dialogue histories and responses. CGDIALOG randomly selected dialogues from the ESConv~\cite{liu-etal-2021-towards} and MSC~\cite{xu-etal-2022-beyond} datasets and constructed 694 history-response pairs for ESConv and 800 history-response pairs for MSC. 
Due to the relatively small size of CGDIALOG, we extend it to CGDIALOG+ by adding 950 history-response pairs from the dialogues in DREAM~\cite{sun-etal-2019-dream}, following the same annotation instruction of CGDIALOG. In the first round of annotation, we hire four graduate computer science students to annotate causal graphs. Subsequently, in the second round, we select annotators who have high-quality annotation results to review all annotations and correct mistakes. We measure the inter-annotator agreement at both the utterance level and the clause level. At the utterance level, we compute Cohen's Kappa and obtain $0.8021$. At the clause level, we compute the averaged F1 score for all possible pairs of annotators and obtain an F1 score of $0.8316$. Both utterance and clause level scores indicate a high level of inter-annotator agreement. The statistics of CGDIALOG+ can be found in Table~\ref{tab:cgdialog_statistics}. More details of data annotation are presented in Appendix~\ref{apx:annotation of cgdialog+}.

\begin{table}[h]
\centering
\resizebox{\linewidth}{!}{%
\begin{tabular}{ l|lll }
\hline 
\textbf{Number of items}                                         & \textbf{ESConv}             & \textbf{MSC}                & \textbf{DREAM }             \\  \hline 
History-response pairs                                  & 694                & 800                & 950                \\ 
Utterances                                              & 2301               & 3807               & 3862               \\ 
\begin{tabular}[c]{@{}l@{}}Direct causes utterance\end{tabular}                 & 1347                & 1525                & 1519               \\ 
\begin{tabular}[c]{@{}l@{}}Average length \\ of direct causes\end{tabular}               & \begin{tabular}[c]{@{}l@{}}24.01\\($\sigma=16.61$)\end{tabular} & \begin{tabular}[c]{@{}l@{}}22.22\\($\sigma=13.79$)\end{tabular} & \begin{tabular}[c]{@{}l@{}}16.67\\($\sigma=11.83$)\end{tabular} \\ 
\begin{tabular}[c]{@{}l@{}}Percentage of causes \\ in their utterances \end{tabular} & \begin{tabular}[c]{@{}l@{}}0.86\\($\sigma=0.22$)\end{tabular}   & \begin{tabular}[c]{@{}l@{}}0.72\\($\sigma=0.27$)\end{tabular}   & \begin{tabular}[c]{@{}l@{}}0.82\\($\sigma=0.20$)\end{tabular}   \\ 
\hline
\end{tabular}
}
\caption{Statistics of the CGDIALOG+.}
\label{tab:cgdialog_statistics}
\vspace{-4mm}
\end{table}

\subsection{Construction of Classifiers}
To construct classifiers, we assume there is a projection function $g(c_i)=\vz_i$, which maps an utterance to a \emph{continuous} latent random variable $\vz_i$ denoting the meaning of the utterance; the corresponding node in the causal graph is denoted by $v_i$. Utterances with similar meaning are thus mapped to the same latent representation. Thus, we are able to build a classifier on top of the hidden representations produced by a pre-trained encoder, e.g. \roberta~\cite{liu2020roberta}.
%


\paragraph{Unconditional Independence Classifier}
The input of the classifier is an utterance $c_i$ and a response $r_t$. The classifier predicts such a pair as positive ($l=1$) if $c_{i} \dep r_t$, otherwise negative ($l=0$). 



To construct a training set, we label a pair of $(c_i, r_t)$ as positive, if either they have a causal relation in CGDIALOG+ or $c_i$ is the preceding utterance of $r_t$. This is supported by the study of \citet{feng-etal-2023-less}, which demonstrates that $90\%$ of preceding utterances serve as direct causes of the following responses. We obtain negative examples by randomly sampling utterances as responses from other conversations.

%
%


We use \roberta as the backbone model to develop the unconditional independence classifier. This is done by integrating a binary classification head, which is fed by the representation of the [CLS] token. As input to \roberta, we concatenate the context utterance $c_{i}$ with the response $r_t$ using the special token '</s>' as the delimiter. This amounts to the unconditional classifier $\mC_{uncond}$.

\paragraph{Conditional Independence Classifier} The input to the CI classifier is the concatenation of two utterances from a dialogue history and a response. It predicts positive if they are conditionally dependent, otherwise negative.

The construction of the initial training set is based on CGDIALOG+. Given one history-response pair from CGDIALOG+, we select one annotated cause of response, one utterance that is unconditionally dependent on the response (determined by $\mC_{uncond}$), and the response as the positive example. Negative examples are constructed similarly but with a crucial difference: instead of using the cause of response, we choose an utterance that is not the cause of response. The constructed dataset is denoted as $\sD_{L}$. We use incremental self-training with constraints to improve the performance of the CI classifier. This method starts with the supervised training of an initial classifier $\mC_0$ on $\sD_{L}$. Then, $\mC_0$ is applied to unlabeled utterance tuples. Those tuples classified with a label of $1$ are incorporated into the training set as positive examples if they satisfy two criteria: 1) the probability $p(l=1| c_i, c_k, r_t)$ surpasses a predefined threshold $0.9$; 2) $c_i$ is $c_{t-2}$ or $c_{t-3}$. Then, a new classifier $\mC_1$ is trained on the updated training set $\sD^{0}$. The self-training cycle is repeated, each iteration yielding a new classifier $\mC_i$, until optimal performance is achieved on the validation set. The classifier ultimately chosen through this self-training process is denoted as $\mC_{cond}$. More details of training the CI classifier are provided in Algorithm~\ref{alg:train_CIC}.

\subsection{Compute \ourmethod}
We compute \ourmethod of responses by using the (un)conditional independence classifiers. Given a response, we first identify individual utterances $c_i$ that have a probability $P(l=1 | c_i, r_t)$ over 0.5 as dependent utterances using the unconditional independence classifier. The set of dependent utterances $c_i$ is denoted by $\gU^{dep}$. Each of those utterances is paired with another utterance in $\gU^{dep}$ and the response to compute the probability of being conditionally dependent. 
The total causal strength of a response w.r.t. a dialogue history is averaged across the corresponding classifier predictions detailed below. 


\citet{138734f0-eb1f-3daf-a1d4-a16462fb1be7, geiger2014estimating} shows causal strength between two variables, $\gS_{v_i\rightarrow v_j}$, can be measured by (C)MI $I(v_i; v_j)$ or $I(v_i;v_j|PA_{v_j}^{-v_i})$ in different causal relations \footnote{\citet{138734f0-eb1f-3daf-a1d4-a16462fb1be7} shows $\gS_{v_i\rightarrow v_j}= I(v_i;v_j)$ or $\gS_{v_i\rightarrow v_j}\geq I(v_i;v_j|PA_{v_j}^{-v_i})$ in different causal relations.}, where $PA_{v_j}^{-v_i}$ represent parents of $v_j$ excluding $v_i$. Considering the diversity and complexity of causal relations in dialogues, we employ both $I(v_i;v_j)$ and $I(v_i;v_j|PA_{v_j}^{-v_i})$ to measure causal strength. It makes sense because both $I(v_i;v_j)$ and $I(v_i;v_j|PA_{v_j}^{-v_i})$ measure strength of dependencies, and strength of dependencies imply causal strength~\cite{138734f0-eb1f-3daf-a1d4-a16462fb1be7}. However, it is still challenging to compute MI or CMI in the dialogue scenario. Considering the equivalent relation between CI test and CMI, we use the probabilities of being dependent or conditional dependent produced by CI classifiers to measure causal strength.

Specifically, the unconditional independence classifier $\mC_{uncond}$ is applied to each pair of $(c_{i}, r_t)$, where $c_{i}$ is an utterance in $\gU^{dep}$. Then, we assess the unconditional dependence strength between each utterance and the response using probability $P(l=1 | c_i, r_t)$, where label 1 represents dependence. We denote this probability as $p_{+}(c_i,r_t)$ for simplicity. 

The conditional classifier $\mC_{cond}$ is thus employed on tuples of the form $(c_i, c_j, r_t)$, where both $c_i$ and $c_j$ are members of the set $\gU_{dep}$ with $i\neq j$. We then compute the probability of $P(l=1 | c_i, c_j, r_t)$ to assess the strength of conditional dependence between utterance and response, $p_{+}(c_i, c_j, r_t)$ for simplicity.
The scoring mechanism for \ourmethod considers both $p_{+}(c_i, r_t)$ and $p_{+}(c_i, c_j, r_t)$ as follows:

\vspace{-4mm}
\begin{equation}
\small
\begin{split}
    &\ourmethod(\vc, r_t) =\\
    &\frac{1}{2} \left ( \frac{\sum_{\gU_{c_i}^{dep}}p_{+}(c_i,r_t)}{ \left | \gU^{dep}\right |} + \frac{\sum_{\gU_{c_i, c_j}^{dep}} p_{+}(c_i, c_j, r_t))}{\left |\gU_{c_i, c_j}^{dep}\right |} \right )
\end{split}
\vspace{-3mm}
\end{equation}
where $c_i$ and $c_j$ are elements of the set $\gU^{dep}$. $\gU_{c_i}^{dep}$ represents select one element from $\gU^{dep}$.  $\gU_{c_i, c_j}^{dep}$ represents select two different elements from $\gU^{dep}$. $\left |\gU_{c_i, c_j}^{dep}\right |$ represents the number of all possible pairs of $(c_i, c_j)$. The score of \ourmethod ranges from 0 to 1, with higher values indicating better relevance.

\section{Experiments}
\subsection{Experimental Setup}
\paragraph{Baseline Metrics.}
We compare our metric \ourmethod with eight dialogue evaluation metrics, consisting of five reference-based metrics:  BLEU~\cite{papineni-etal-2002-bleu}, ROUGE~\cite{lin-2004-rouge}, METEOR~\cite{lavie-agarwal-2007-meteor}, BERTScore~\cite{BERTScore}, BLEURT~\cite{sellam-etal-2020-bleurt}. For comparison, we only present the BLEU-4 for BLEU, ROUGE-L for ROUGE, and BERTScore-F1 for BERTScore. Based on the prior works~\cite{li2022diffusionlm, yang-klein-2021-fudge, Dathathri2020Plug}, we feed the dialogue history and corresponding generated text to a language model (\ie a carefully fine-tuned GPT-2 model) and report the perplexity (PPL) of the generated text under the language model. GRADE~\cite{huang-etal-2020-grade} and DEAM~\cite{ghazarian-etal-2022-deam} evaluate dialogues by using probability of fine-tuned classifiers. DEnsity~\cite{park-etal-2023-density} evaluates a response by utilizing density estimation on the feature space derived from a neural classifier. To ensure a fair comparison, these classifier-based models are fine-tuned on experimental dialogue datasets. Because \citet{chiang-lee-2023-large, wang2023chatgpt} argue that ChatGPT can be a good text generation evaluation metric, we also consider ChatGPT as a baseline metric for dialogue evaluation. We follow the prompts from \citet{chiang-lee-2023-large, wang2023chatgpt} to require ChatGPT to evaluate responses using a 5-point Likert scale.

\paragraph{Datasets.}
We conduct experiments on three dialogue datasets across diverse domains: ESConv~\cite{liu-etal-2021-towards}, MSC~\cite{xu-etal-2022-beyond}, DREAM~\cite{sun-etal-2019-dream}. The details of the datasets are provided in Appendix~\ref{apx:datasets}. For MSC and DREAM, we use the dataset splits as provided in their publications. For ESConv, because it doesn’t have an official split, we randomly split the dataset with $80\%$ dialogues for training, $10\%$ dialogues for validation, and $10\%$ for testing. As a result, any dialogue in a test set cannot be seen in any of the training sets.


\paragraph{Implementation Details.}
We use RoBERTa~\cite{liu2020roberta} as the backbone model to fine-tune classifiers. All the models are implemented with PyTorch~\cite{paszke2019pytorch} and the Transformers library~\cite{wolf-etal-2020-transformers}. All models are trained with Adam~\cite{kingma2017adam} optimizer with $\beta_{1}=0.9$ and $\beta_{2}=0.999$. The learning rate is $1\times 10^{-5}$ for fine-tuning classifiers. We use a linear learning rate scheduler that dynamically decreases the learning rate after 10 warm-up steps. Classifiers were trained for $10$ epochs with the batch size $16$ on NVIDIA A40 GPU.

\paragraph{Dialogue Models.} We evaluate metrics using both human-generated and model-generated responses to assess their performance across varying levels of response quality. For model-generated responses, we consider two dialogue models, \blenderbot~\cite{roller-etal-2021-recipes} and \blenderbot-\constrain~\cite{feng-etal-2023-less}, both known for producing human-like responses. Additionally, we fine-tuned a large language model named \alpaca~\cite{alpaca} using the LoRA technique~\cite{hu2022lora} on dialogue datasets.

\subsection{Metric Evaluation}

\paragraph{Human Judgements.}
\citet{belz-kow-2010-comparing, callison-burch-etal-2007-meta, Kiritchenko2017BestWorstSM} found that asking crowd-workers to directly score responses on a Likert scale usually receives low-quality evaluation. Thus, following the evaluation design in \citet{novikova2018rankme, bojar-etal-2016-findings, zheng-etal-2021-comae, zhou2018emotional, feng-etal-2023-less}, we opt for pairwise comparison between responses from different dialogue models. For each dataset, we randomly sample 100 dialogue histories from the test set. Then, given one dialogue history, we ask annotators to compare two responses from two dialogue models. Because we have four dialogue models (one human response and three model-generated responses), there are six different pair comparisons in total. The annotation was conducted by 16 undergraduate and graduate students who are native English speakers.

In each comparison, we ask five evaluation questions: \textbf{Empathy} (Which response has a better understanding of the emotional state and provides a more appropriate emotional reaction?), \textbf{Specificity} (Which response produces more unique and non-generic information that is specific to the conversation history?), \textbf{Relevance} (Which response is more on-topic with the immediate dialogue history?), \textbf{Consistency} (Which response is more logically coherent with the conversation history and common sense?) and \textbf{Overall} (Which response performs better overall?). Each question has four options: \textit{A is better than B}, \textit{B is better than A}, \textit{Both are good}, and \textit{Both are bad}. Three individual annotators assessed each comparison. To eliminate any bias from annotators, we anonymized the names of dialogue models, shuffled the order of dialogues, and shuffled the order of responses. Finally, we collected 1800 pairwise comparison results from 16 annotators. The calculated Krippendorff's alpha~\cite{Krippendorff2011ComputingKA} for assessing the inter-annotator agreement is 0.6708, indicating a moderate level of agreement among the annotators.

\paragraph{Correlation Calculation.}
Because human evaluation results are categorical options and automatic metrics are continuous values, we cannot directly calculate correlation coefficients between them. Thus, we apply different schemas to convert categorical options to integer values and convert continuous values to categorical options. 

To convert categorical options into integer values, we use the \textbf{voting schema}. Specifically, if one annotator selects \textit{A is better than B}, response A gets one point, while B gets zero points, and vice versa for \textit{B is better than A}. If one annotator selects \textit{Both are good}, both responses A and B get one point. If \textit{Both are bad} is selected, both responses A and B get zero points. Then we apply this rule to three annotator assessments. After conversion, we have integer scores for human evaluation and continuous scores for automatic evaluation. Then, we apply Pearson and Spearman's correlation coefficient to measure correlations between human evaluation and automatic evaluation. Because continuous metrics hardly produce exactly equivalent values, we propose a \textbf{IgnoreEqual} schema that only considers nonequivalent relationships. Specifically, for one human annotator results, we convert \textit{A is better than B} to 1 and \textit{A is better than B} to 0. In this way, human evaluation becomes a dichotomous variable. For automatic evaluation, we consider the difference of automatic score on response A and response B. Formally, we take $AutoMetric(A)-AutoMetric(B)$ as another variable, where $AutoMetric$ refers to any automatic metric, $A$ and $B$ refer to response A and response B. In this way, we can use Point-Biserial correlation coefficient to correlation between human evaluation and automatic evaluation.
To convert continuous values into category options (\textbf{Cont2Cat}), we simply compare automatic scores of responses A and B. If the score of response A is larger than B, we convert it to \textit{A is better than B}, otherwise convert it to \textit{B is better than A}. After conversion, we treat automatic metric as another annotator and compute inter-annotator agreement using Krippendorff's alpha method. 

\subsection{Analytical Experiments}

\begin{table*}[h]
\centering
\resizebox{\textwidth}{!}{%
\begin{tabular}{r|cccccccccccc}
\hline
\multicolumn{1}{c|}{}                        & \multicolumn{4}{c|}{\textbf{DREAM}}                                                        & \multicolumn{4}{c|}{\textbf{ESConv}}                                                      & \multicolumn{4}{c}{\textbf{MSC}}                                     \\ \hline
\multicolumn{1}{c|}{\multirow{2}{*}{Metric}} & \multicolumn{2}{c}{Voting}        & IgnoreEqual     & \multicolumn{1}{c|}{Cont2Cat}        & \multicolumn{2}{c}{Voting}        & IgnoreEqual     & \multicolumn{1}{c|}{Cont2Cat}       & \multicolumn{2}{c}{Voting}        & IgnoreEqual     & Cont2Cat       \\
\multicolumn{1}{c|}{}                        & Pearson         & Spearman        & Point-Biserial  & \multicolumn{1}{c|}{IAA}             & Pearson         & Spearman        & Point-Biserial  & \multicolumn{1}{c|}{IAA}            & Pearson         & Spearman        & Point-Biserial  & IAA            \\ \hline
\multicolumn{1}{l|}{}                        & \multicolumn{12}{c}{\textbf{Relevance}}                                                                                                                                                                                                                       \\ \hline
\multicolumn{1}{l|}{\ourmethod}              & \textbf{0.294*} & \textbf{0.334*} & \textbf{0.363}  & \multicolumn{1}{c|}{\textbf{0.369}}  & \textbf{0.312*} & \textbf{0.343*} & \textbf{0.402}  & \multicolumn{1}{c|}{\textbf{0.337}} & \textbf{0.257*} & \textbf{0.308*} & \textbf{0.316}  & \textbf{0.330} \\
-$p(c_i, c_j, r_t)$                          & 0.184           & 0.157           & 0.216           & \multicolumn{1}{c|}{0.312}           & 0.148           & 0.146           & 0.209           & \multicolumn{1}{c|}{0.284}          & 0.137           & 0.151           & 0.176           & 0.289          \\
-$p(c_i, r_t)$                               & 0.229           & 0.303*          & 0.335*          & \multicolumn{1}{c|}{0.347}           & 0.294*          & 0.328*          & 0.362           & \multicolumn{1}{c|}{0.327}          & 0.204           & 0.256*          & 0.292           & 0.318          \\
-self-training                               & 0.285*          & 0.325*          & 0.351*          & \multicolumn{1}{c|}{0.358}           & 0.302*          & 0.340*          & 0.387*          & \multicolumn{1}{c|}{0.336}          & 0.247*          & 0.299*          & 0.304*          & 0.324          \\
$\rightarrow$ MaxCI                               & 0.087          & 0.075          & 0.101          & \multicolumn{1}{c|}{0.302}           & 0.095          & 0.079          & 0.104          & \multicolumn{1}{c|}{0.277}          & 0.133          & 0.119          & 0.161          & 0.271          \\  
$\rightarrow$ Preced2                               & 0.150          & 0.128          & 0.177          & \multicolumn{1}{c|}{0.303}           & 0.114          & 0.107          & 0.163          & \multicolumn{1}{c|}{0.280}          & 0.105          & 0.121          & 0.146          & 0.272          \\
\hline
\multicolumn{1}{l|}{}                        & \multicolumn{12}{c}{\textbf{Overall}}                                                                                                                                                                                                                         \\ \hline
\multicolumn{1}{l|}{\ourmethod}              & \textbf{0.331*} & \textbf{0.422*} & \textbf{0.511*} & \multicolumn{1}{c|}{\textbf{0.595}}  & \textbf{0.287*} & \textbf{0.339*} & \textbf{0.411*} & \multicolumn{1}{c|}{\textbf{0.568}} & \textbf{0.331*} & \textbf{0.401*} & \textbf{0.492*} & \textbf{0.569} \\
-$p(c_i, c_j, r_t)$                          & 0.192           & 0.231           & 0.303           & \multicolumn{1}{c|}{0.517}           & 0.115           & 0.121           & 0.161           & \multicolumn{1}{c|}{0.483}          & 0.179           & 0.235           & 0.272           & 0.526          \\
-$p(c_i, r_t)$                               & 0.303*          & 0.396*          & 0.496*          & \multicolumn{1}{c|}{0.571}           & 0.262*          & 0.314*          & 0.403*          & \multicolumn{1}{c|}{0.548}          & 0.316*          & 0.380*          & 0.473*          & 0.546          \\
-self-training                               & 0.326*          & 0.414*          & 0.503*          & \multicolumn{1}{c|}{0.586}           & 0.284*          & 0.331*          & 0.407*          & \multicolumn{1}{c|}{\textbf{0.568}} & 0.324*          & 0.387*          & 0.488*          & 0.562 \\
$\rightarrow$ MaxCI                               & 0.203          &   0.147         & 0.250          & \multicolumn{1}{c|}{0.490}           & 0.048          & 0.087          & 0.058          & \multicolumn{1}{c|}{0.473}          & 0.086          & 0.116          & 0.112 & 0.480                \\ 
$\rightarrow$ Preced2                               & 0.172          & 0.158          & 0.183          & \multicolumn{1}{c|}{0.358}           & 0.103          & 0.095          & 0.135          & \multicolumn{1}{c|}{0.263}          & 0.126          & 0.131          & 0.157          & 0.301          \\
\hline
\end{tabular}%
}
\caption{Ablation results on three datasets. Asterisk * indicates results with p-value < 0.05 (statistically significant).}
\label{tab:ablation_study_corr_partial}

\vspace{-4mm}
\end{table*}
To comprehensively evaluate the individual contributions of \ourmethod, we conducted a series of ablation studies. The outcomes of these studies are presented in Table~\ref{tab:ablation_study_corr_partial}. The observed decline in the removal of each component demonstrates their collective positive impact on the evaluation of responses, thus supporting the integral role of each element within the \ourmethod framework.

\paragraph{Efficacy of the Classifiers.} To prove the contribution of (conditional) mutual information in our framework, we perform two ablation experiments: 1) removing unconditional dependence (\ie -$p(c_i, r_t)$ rows of Table~\ref{tab:ablation_study_corr_partial}) in the computation of \ourmethod scores; 2) removing conditional dependence (\ie -$p(c_i, c_j, r_t)$ rows) when computing \ourmethod scores. Consequently, removing conditional dependence has the most detrimental impact on the metric's performance. As we described in Section~\ref{sec:methodology}, we argue that a response exhibiting high relevance to the dialogue history inherently entails strong causal strength with a particular set of utterances in that dialogue history. Furthermore, causal strength can be well measured by the degree of conditional independence. In other words, conditional dependence is closer to causal strength than unconditional dependence.  The better performance of CI classifier can be attributed to the fact that conditional dependencies more accurately reflect the actual causal relations between the dialogue history and the response than unconditional dependencies. 

Instead of taking the average of conditional dependence, we only use the maximum of conditional dependence to compute the metric score as another ablation study (\ie $\rightarrow$ MaxCI rows). In several instances, relying on the maximum conditional dependence yields inferior results compared to using the average of unconditional dependencies. This outcome can be attributed to the fact that the relevance of responses is more accurately reflected by the causal relations with the entire dialogue history, rather than only with the most likely direct cause.

\paragraph{Usefulness of Annotated Causal Relations.}
We verify the necessity of annotated causal relations on training the CI classifier. Instead of using annotated causal relations, we trained a CI classifier using the two most recent preceding utterances as positive instances and two random utterances from other dialogue as negative instances. The performance outcomes, detailed in the "-Preced2" row, demonstrate a notable decline when compared to the CI classifier trained on the annotated CGDIALOG+ dataset (\ie -$p(c_i, r_t)$). We attribute this performance drop to the high noise present in the positive examples. Our empirical observations suggest that approximately only 40\% of the penultimate utterances serve as the cause of response, indicating a significant level of noise within positive instances, which undermines the classifier’s reliability.

\paragraph{Effectiveness of Self-Training.} We compare the CI classifier $\mC_{cond}$ trained with incremental self-training with constraints with the initial classifier $\mC_{0}$. As shown in the '-self-training' rows of Table~\ref{tab:ablation_study_corr_partial}, \ourmethod without self-training results in a decline of 0.012 in Pearson correlation, 0.013 in Spearman correlation, 0.007 in Point-Biserial correlation, 0.012 in inter-annotator agreement in average. We believe self-training with constraints benefits the training of CI classifiers by augmenting training data and reducing the noise in pseudo-label data. These findings indicate incremental self-training with constraints is an effective method to improve the performance of classifiers. 

\subsection{Experimental Results}
\label{sec:results}
\begin{table*}[!h]
\centering
\resizebox{\textwidth}{!}{%
\begin{tabular}{ll|cccc|cccc|cccc}
\hline
\multicolumn{2}{c|}{\textbf{}}       & \multicolumn{4}{c|}{\textbf{DREAM}}                                 & \multicolumn{4}{c|}{\textbf{ESConv}}                                & \multicolumn{4}{c}{\textbf{MSC}}                                    \\ \hline
\multicolumn{2}{c|}{\multirow{2}{*}{Metric}}  & \multicolumn{2}{c}{Voting}        & IgnoreEqual    & Cont2Cat       & \multicolumn{2}{c}{Voting}        & IgnoreEqual    & Cont2Cat       & \multicolumn{2}{c}{Voting}        & IgnoreEqual    & Cont2Cat       \\
\multicolumn{2}{c|}{}                         & Pearson         & Spearman        & Point-Biserial & IAA            & Pearson         & Spearman        & Point-Biserial & IAA            & Pearson         & Spearman        & Point-Biserial & IAA            \\ \hline
\multicolumn{14}{c}{\textbf{Relevance}}  \\ \hline
\multirow{5}{*}{Reference-based} & BLEU       & 0.021           & 0.018           & 0.027          & 0.246          & -0.047          & -0.065          & -0.053         & 0.216          & 0.076           & 0.065           & 0.074          & 0.222          \\
                                 & ROUGE      & -0.005          & -0.008          & -0.015         & 0.272          & 0.039           & 0.020           & 0.045          & 0.243          & 0.091           & 0.087           & 0.090          & 0.249          \\
                                 & METEOR     & 0.028           & 0.033           & 0.043          & 0.262          & 0.097           & 0.097           & 0.081          & 0.243          & 0.013           & 0.036           & 0.087          & 0.241          \\
                                 & BERTScore  & -0.004          & -0.010          & -0.003         & 0.260          & 0.085           & 0.069           & 0.092          & 0.246          & 0.021           & 0.002           & 0.031          & 0.239          \\
                                 & BLEURT     & -0.022          & -0.032          & -0.030         & 0.257          & 0.025           & 0.022           & 0.026          & 0.246          & 0.074           & 0.089           & 0.094          & 0.249          \\ \hline
\multirow{5}{*}{Reference-free} & PPL      & 0.033           & 0.097          & 0.043          & 0.292          & -0.040           & -0.031           & -0.073          & 0.246          & -0.046          & -0.047           & -0.053          & 0.245          \\ 
                                & GRADE      & 0.004           & -0.005          & 0.035          & 0.248          & 0.013           & 0.021           & 0.030          & 0.248          & -0.003          & 0.012           & 0.049          & 0.243          \\
                                 & DEAM       & -0.090          & -0.053          & -0.121         & 0.273          & -0.011          & 0.039           & -0.011         & 0.257          & -0.012          & -0.032          & -0.007         & 0.253          \\
                                 & DEnsity       &  0.117         &  0.112         &  0.149        &  0.286         &  0.080         &   0.099          & 0.095         & 0.268          &  0.030         & 0.030          &  0.026        &  0.258         \\
                                 & ChatGPT    & 0.036           & 0.024           & 0.088          & 0.284          & -0.002          & -0.018          & 0.096          & 0.250          & 0.083           & 0.084           & 0.109          & 0.271          \\
                                 & GPT4       & 0.049           & 0.038           & 0.083          & 0.263          & -0.002          & -0.023          & 0.097          & 0.251          & 0.039           & 0.083           & 0.110          & 0.277          \\
                                 & \ourmethod & \textbf{0.294*} & \textbf{0.334*} & \textbf{0.363} & \textbf{0.369} & \textbf{0.312*} & \textbf{0.343*} & \textbf{0.402} & \textbf{0.337} & \textbf{0.257*} & \textbf{0.308*} & \textbf{0.316} & \textbf{0.330} \\ \hline \hline
\multicolumn{14}{c}{\textbf{Overall}}  \\ \hline
\multirow{5}{*}{Reference-based} & BLEU       & 0.019           & 0.058           & 0.011           & 0.444          & 0.069           & 0.050           & 0.005           & 0.434          & 0.019           & -0.019          & 0.007           & 0.422          \\
                                 & ROUGE      & -0.031          & -0.028          & -0.040          & 0.453          & -0.030          & -0.041          & -0.044          & 0.445          & -0.011          & -0.018          & -0.010          & 0.435          \\
                                 & METEOR     & -0.043          & -0.031          & -0.053          & 0.454          & 0.041           & 0.029           & 0.052           & 0.455          & 0.006           & 0.015           & 0.007           & 0.435          \\
                                 & BERTScore  & 0.065           & 0.077           & 0.103           & 0.458          & -0.028          & 0.004           & -0.035          & 0.458          & 0.032           & 0.042           & 0.052           & 0.440          \\
                                 & BLEURT     & 0.011           & 0.005           & 0.011           & 0.451          & -0.112          & -0.117          & -0.161          & 0.439          & 0.076           & 0.077           & 0.112           & 0.445          \\ \hline
\multirow{5}{*}{Reference-free} & PPL      & 0.034           & 0.010           & 0.032           & 0.454          & 0.045           & 0.105           & 0.100           & 0.480          & 0.023           & -0.038           & -0.022           & 0.436          \\ 
                                & GRADE      & 0.054           & 0.033           & 0.012           & 0.454          & 0.023           & 0.011           & 0.004           & 0.436          & 0.088           & 0.050           & 0.105           & 0.442          \\
                                 & DEAM       & 0.111           & 0.107           & 0.168           & 0.467          & 0.013           & 0.010           & 0.005           & 0.442          & 0.042           & 0.021           & 0.074           & 0.442          \\
                                 & DEnsity       &  0.011         &  0.009         &  0.023        &  0.462         &  0.038         &    0.100          & 0.064         & 0.483          &  0.076         & 0.045          &  0.091        &  0.465         \\
                                 & ChatGPT    & 0.153           & 0.101           & 0.113           & 0.460          & 0.052           & 0.055           & 0.041           & 0.463          & 0.129           & 0.125           & 0.181           & 0.481          \\
                                 & GPT4       & 0.159           & 0.157           & 0.141           & 0.486          & 0.048           & 0.062           & 0.042           & 0.471          & 0.131           & 0.103           & 0.119           & 0.486          \\
                                 & \ourmethod & \textbf{0.331*} & \textbf{0.422*} & \textbf{0.511*} & \textbf{0.595} & \textbf{0.287*} & \textbf{0.339*} & \textbf{0.411*} & \textbf{0.568} & \textbf{0.331*} & \textbf{0.401*} & \textbf{0.492*} & \textbf{0.569} \\ \hline
\end{tabular}%
}
\caption{Correlations between automatic evaluation metrics and human judgements on three different datasets (DREAM, ESConv, MSC). Inter-annotator agreement (IAA) is computed using Krippendorff's alpha. PPL represents perplexity. Asterisk * indicates results with p-value < 0.05 (statistically significant).}
\label{tab:metric_corr_all_dims_partial}
\end{table*} 

Table~\ref{tab:metric_corr_all_dims_partial} depicts the quantitative results for different evaluation metrics on ESConv, MSC, and DREAM datasets. According to the reported correlations and inter-annotator agreements, \ourmethod outperforms all baseline metrics across various evaluation dimensions, including relevance, specificity, empathy, consistency, and overall. \ourmethod achieves higher correlations on relevance which is the primary target evaluation dimension of our metric. Regarding the overall dimension, it is posited that annotators tend to favor responses having high relevance, perceiving them as indicative of superior overall quality. This comprehensive effectiveness of \ourmethod can be ascribed to its capability to identify causal relations between dialogue histories and responses. Such causal connections are essential to establish the relevance of responses in the context of the preceding dialogue.

Baseline metrics can be categorized into two types: reference-based and reference-free metrics. Our experimental findings reveal that both types are generally unreliable for evaluating responses. Although ChatGPT and GPT-4-based metrics exhibit relatively better correlations in the dimensions of empathy and consistency, this enhanced performance lacks stability across different datasets.

\subsection{Qualitative Study}
\begin{table*}[h]
\centering
\resizebox{\textwidth}{!}{%
\begin{tabular}{llccccc}
\hline
            & \multicolumn{6}{l}{……}                                                                                                                                                                                                                                                              \\
M:          & \multicolumn{6}{l}{Good afternoon. I've just seen the advertisement in the newspaper about the furnished flat for rent. Is it still available?}                                                                                                                                     \\
W:          & \multicolumn{6}{l}{Yes, it is. Would you like me to tell you about it?}                                                                                                                                                                                                             \\
M:          & \multicolumn{6}{l}{Yes, but could you tell me the address first, please?}                                                                                                                                                                                                           \\
W:          & \multicolumn{6}{l}{Yes, it's Number 45 Station Road.}                                                                                                                                                                                                                               \\
M:          & \multicolumn{6}{l}{Thank you. How big is the flat?}                                                                                                                                                                                                                                 \\
W:          & \multicolumn{6}{l}{Oh, it's big enough for a family of four. There are three bedrooms.}                                                                                                                                                                                             \\
M:          & \multicolumn{6}{l}{Thanks. What about the heating and the stove?}                                                                                                                                                                                                                   \\
W:          & \multicolumn{6}{l}{It's all gas -- the flat has central heating and a gas stove.}                                                                                                                                                                                                   \\
M:          & \multicolumn{6}{l}{OK. And how much is the rent?}                                                                                                                                                                                                                                   \\ \hline
            & \multicolumn{1}{l|}{}                                                                                                                                                                                                                 & \multicolumn{5}{c}{Evaluation on Relevance} \\
Responses:  & \multicolumn{1}{l|}{}                                                                                                                                                                                                                 & Human  & \ourmethod  & GPT4  & DEAM  & BERTScore \\ \hline
Human:      & \multicolumn{1}{l|}{\begin{tabular}[c]{@{}l@{}}It's \$ 200 a month, and that does not include the cost of gas. \\ The flat will be available starting Sep. 1.\end{tabular}}                                                            & 9      & 0.611       & 5     & 0.996 & 1.0  \\
\alpaca:     & \multicolumn{1}{l|}{\begin{tabular}[c]{@{}l@{}}The flat is located at Number 45 Station Road and is big enough for a family of four. \\ It has three bedrooms and is gas heated with a gas stove. The rent is \$6487258.\end{tabular}} & 7      & 0.583       & 1     & 0.995 & 0.843  \\
\blenderbot: & \multicolumn{1}{l|}{I don't know yet, but it's a lot of money.}                                                                                                                                                                       & 8      & 0.602       & 1     & 0.996 & 0.855  \\
\constrain:  & \multicolumn{1}{l|}{It looks nice. Would you like to see it?}                                                                                                                                                                         & 2      & 0.557       & 1     & 0.996 & 0.846  \\ \hline
\end{tabular}%
}
\caption{A case study showing evaluation results of human judgement, \ourmethod, GPT4, DEAM, and BERTScore. We use voting schema on all pairwise comparisons to get human scores. \alpaca represents \alpaca fine-tuned using LoRA. \constrain represents the \blenderbot-\constrain dialogue model. }
\label{tab:case_study}
\vspace{-15px}
\end{table*}
To provide a more intuitive assessment of \ourmethod's performance, we present a representative example in Table~\ref{tab:case_study}. For human evaluations, we can see the human-generated response exhibits the highest relevance. Responses generated by $\alpaca$ and $\blenderbot$ display relatively high relevance, while response generated by $\blenderbot-\constrain$ shows the lowest relevance. Notably, the ranking of scores assigned by our metric aligns more closely with human judgements compared to other metrics. The GPT4-based metric often assigns the highest scores to human responses but falls short of properly ranking model generated responses. The DEAM metric allocates nearly identical scores to all responses, suggesting its inadequacy in differentiating between varying levels of relevance. BERTScore, as a reference-based metric, naturally scores the human response as $1.0$ due to it serves as the reference. However, it assigns similar scores to all model-generated responses, highlighting the inability of reference-based metrics to effectively address the one-to-many nature of dialogues. More examples can be found in Appendix~\ref{apx:qualitative_study}.

\section{Conclusion}
In this paper, we propose \ourmethod, a novel automatic metric for evaluating the relevance of responses. Experimental results show that \ourmethod exhibits stronger correlations with human judgements than the SOTA metrics in terms of diverse correlation measures in each of the datasets. In addition, we release a new dataset CGDIALOG+ annotated with causal relations in dialogues and human preference judgements between responses from various sources. 

\section*{Limitations}
\label{sec:limitations}

Due to the limited budget for this project, we cannot recruit a large number of annotators to annotate large dialogue datasets with causal relations. Consequently, the CGDIALOG+ dataset is relatively modest in size. Although the size of CGDIALOG+ is large enough to conduct experiments to support the key claims of this work, it may not meet the requirements of industrial applications. The industrial users of our dataset may mitigate this issue by using the self-training technique or hiring more annotators to enlarge our datasets following our approach.

Our metric focuses on evaluating the relevance of generated responses. While our metric outperforms the baselines in terms of empathy and consistency, its margin is not as high as in relevance and specificity. Thus, the design of novel metrics for task-specific evaluation criteria will be a promising direction of our future work.

\section*{Ethics Statement}
We acknowledge the importance of ACM Code of Ethics and agree with it. We ensure that our study is compatible with the provided code.

The development and evaluation of \ourmethod have been conducted with a keen awareness of ethical considerations, particularly those pertaining to the use of human annotators. Our approach requires human annotation to construct the training set (CGDIALOG+) for conditional independence classifiers, a process we acknowledge as labor-intensive. We have ensured that the annotation process adheres to ethical guidelines, respecting the time and effort of the annotators and ensuring fair compensation for their contributions. We have taken rigorous measures to anonymize the dataset thoroughly. The dataset does not contain any personally identifiable information or sensitive data related to the contributors. The CGDIALOG+ dataset was compiled with contributions from undergraduate and graduate students, which may inherently introduce biases based on their demographic backgrounds. We advise researchers utilizing this dataset to carefully consider these potential biases, particularly in studies focusing on AI fairness, biases, and safety.

Moreover, the central objective of \ourmethod is to assess the relevance of generated responses in relation to their preceding dialogue history. The metric produces continuous scores that range from 0 to 1, where higher scores denote greater relevance. It is designed to yield only these scores, without generating any information that could be deemed harmful or violate privacy. The development of CausalScore aims to provide a cost-effective alternative to human evaluation, which is both expensive and time-consuming. It is crucial to emphasize that the intention behind CausalScore is not to facilitate the creation of chatbots that engage in unsafe or inconsistent behaviors.

\bibliography{anthology,custom}
\bibliographystyle{acl_natbib}

\appendix

\section{Appendix}
\label{sec:appendix}
\subsection{Related Work}
Automatic evaluation for open-domain dialogue systems is challenging as one dialogue context can have many appropriate responses, which is known as the one-to-many nature of dialogues \citep{zhao-etal-2017-learning, yeh-etal-2021-comprehensive}. In general, dialogue evaluation metrics can be divided into reference-based metrics and reference-free metrics. Reference-based metrics, such as BLEU \citep{papineni-etal-2002-bleu}, ROUGE \citep{lin-2004-rouge}, METEOR\citep{lavie-agarwal-2007-meteor}, BERTScore \citep{BERTScore}, BLEURT \cite{sellam-etal-2020-bleurt}, are widely used for language generation and machine translation tasks. Those metrics use statistical rules or learned embeddings to measure the surface similarity between generated responses and reference responses. However, they cannot deal with the one-to-many nature of dialogues and many works have shown that they have weak correlations with human judgements \citep{huang-etal-2020-grade, yeh-etal-2021-comprehensive, ghazarian-etal-2022-deam, wang2023chatgpt}.

Considering the one-to-many nature of dialogues, recent research has proposed several reference-free automatic metrics, which directly assess generated responses without given references. RUBER proposed by \citet{tao2017ruber} is trained with a triplet ranking loss using an RNN neural network. \citet{huang-etal-2020-grade} propose GRADE metric, which constructs a topic transition graph in dialogues and then feeds the graph and input into a neural network to compute a coherence score. However, GRADE uses commonsense knowledge graph ConceptNet \cite{10.5555/3298023.3298212} to construct topic graphs in dialogues, which may cause wrong assessment due to domain shift. To train reference-free metrics, high-quality incoherent responses are essential. \citet{10.1007/978-3-030-00671-6_37, mesgar-etal-2020-dialogue, zhang-etal-2021-dynaeval} automatically generate incoherent responses by shuffling utterances order, inserting or replacing irrelevant utterances. \citet{ghazarian-etal-2022-deam} relies on abstract meaning representation to apply semantic-level manipulations for incoherent response generation. \citet{chiang-lee-2023-large, wang2023chatgpt} employ large language models (e.g., ChatGPT) as a metric for text generation tasks, utilizing crafted prompts. The experimental results suggest the reliability of using large language models as metrics.

\subsection{Dialogue Datasets}
\label{apx:datasets}
\paragraph{Emotion Support Conversation (ESConv).} ESConv \citep{liu-etal-2021-towards} contains 1,053 conversations between mental health help seekers and supporters, with $29.8$ utterances per dialogue on average. In each dialogue, help seekers talk about their problems, such as unemployment, losing family member or infecting with COVID. Dialogue response models play the role of supporters to provide supportive responses to help seekers. Each utterance from supporters is annotated with a strategy such as providing suggestions, paraphrasing or question, which are not considered in our models. For ESConv, because it doesn’t have an official split, we split dialogues with $80\%$ dialogues for training, $10\%$ dialogues for validation, and $10\%$ for testing.

\paragraph{Multi-Session Chat (MSC).} MSC \citep{xu-etal-2022-beyond} contains 5,000 human-human conversations over five sessions, each of which contains up to 14 utterances. 
The average number of utterances per dialogue is $53.3$. In each session, two interlocutors conduct a conversation based on given personas. Each persona describes personal information with multiple sentences. We use the official split for experiments.

\paragraph{DREAM.} DREAM \citep{sun-etal-2019-dream} collects conversations from English as a Foreign Language examinations designed by human experts to evaluate the comprehension level of Chinese learners of English. It contains $6,444$ dialogues, with $4.7$ utterances per dialogue on average. The topics are about daily life including diverse topics. We use the official split for experiments.

\subsection{Annotation of CGDIALOG+}
\label{apx:annotation of cgdialog+}
We randomly sampled 95 dialogues from DREAM~\cite{sun-etal-2019-dream}, which results in the creation of 950 history-response pairs, annotating about 10 context-response pairs per dialogue. We engaged annotators who have a thorough understanding of identifying direct causes of responses. The annotation process uses Amazon Mechanical Turk (AMT). 

\begin{figure*}
    \centering
    \includegraphics[width=\textwidth]{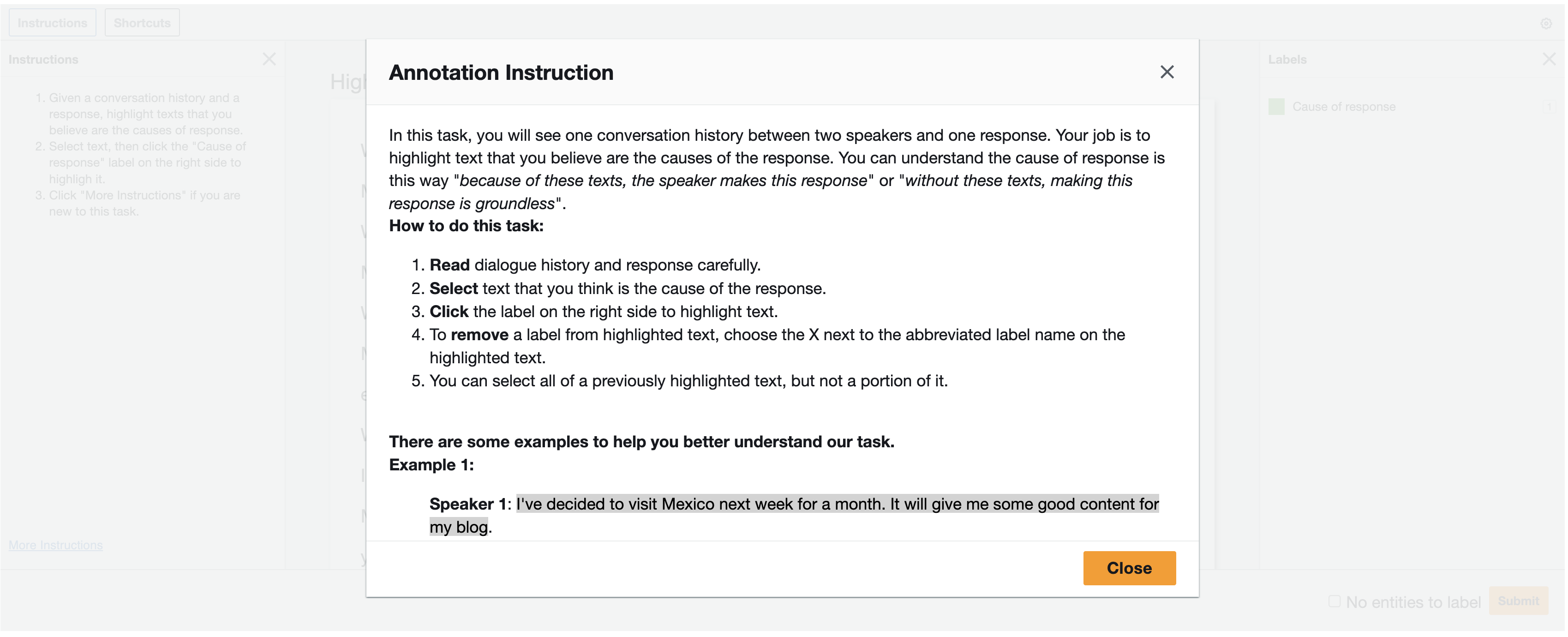}
    \caption{Annotation instruction of CGDIALOG+.}
    \label{fig:annotation_instruction}
\end{figure*}

\begin{figure*}
    \centering
    \includegraphics[width=\textwidth]{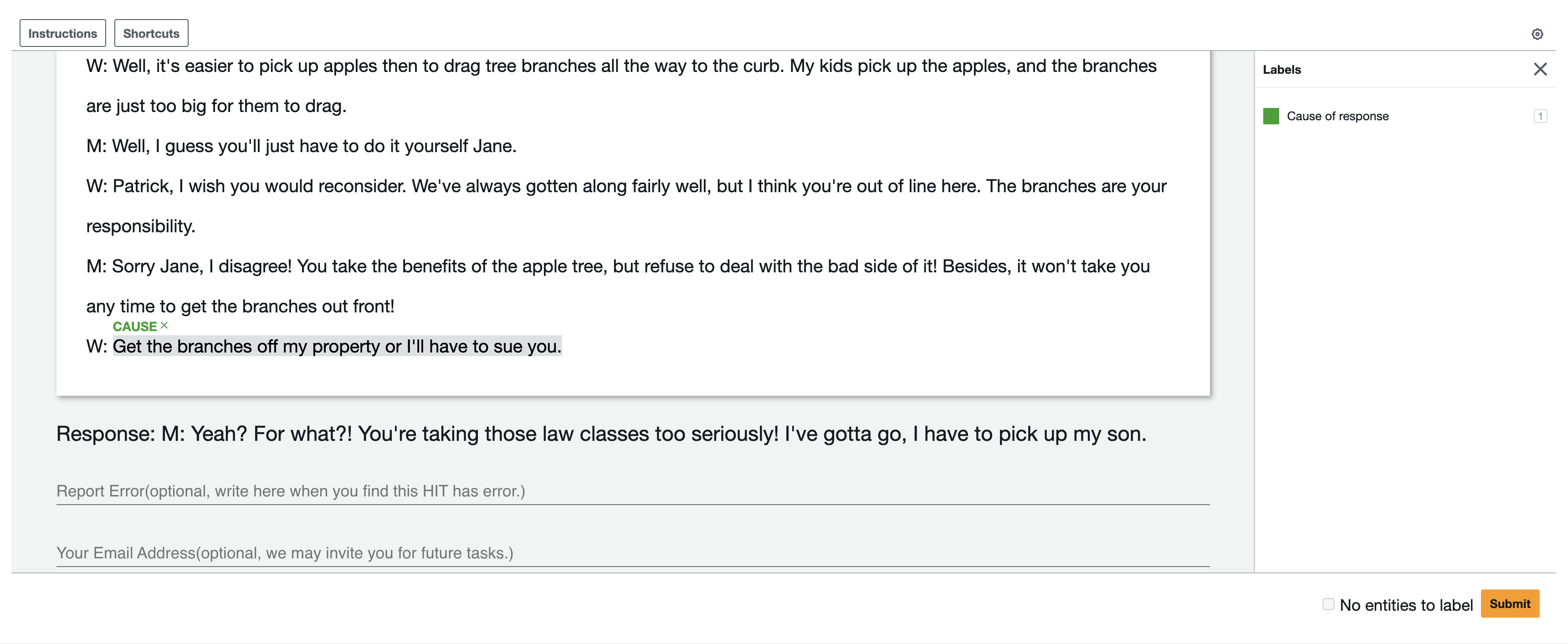}
    \caption{CGDIALOG+ annotation interface.}
    \label{fig:annotation_interface}
\end{figure*}

To ensure the understanding of the task, a training phase was implemented before real annotation. This phase involved a 'dry-run' dataset, where annotators practiced annotation tasks. Comprehensive feedback was provided in cases of any misunderstanding, thereby fine-tuning their annotation skills. After training, in the first annotation round, annotators were asked to read the provided responses and their conversation histories, then highlight utterances or clauses that directly caused the responses. We can understand the cause of response in this way "because of these texts, the speaker makes this response" or "without these texts, making this response is groundless". To maintain high annotation quality, in the second annotation round, we select annotators who have high-quality annotation results to review all annotations and correct mistakes. We carefully distribute the workload among annotators to ensure they do not review their own annotations. Our annotators received compensation exceeding the local minimum hourly wage. Annotation instruction and interface can be found in Figure~\ref{fig:annotation_instruction} and Figure~\ref{fig:annotation_interface}.

For our experimental setup, the CGDIALOG-DREAM dataset was partitioned into a training set comprising 450 context-response pairs, a validation set with 250 pairs, and a test set also containing 250 pairs. The division of the CGDIALOG-ESConv and CGDIALOG-MSC datasets follow their official allocations, which are 272/211/211 and 300/250/250 context-response pairs for training, validation, and testing, respectively.

\subsection{Training of Conditional Independence Classifier}
In Algorithm~\ref{alg:train_CIC}, we provide more details of training the conditional independence classifier.

\begin{algorithm*}
\caption{Training of Conditional Independence Classifier}
\label{alg:train_CIC}
\begin{algorithmic}
\Require 
\State Labeled training and validation sets from CGDIALOG+: $\sD_{L}^{tr}$, $\sD_{L}^{va}$
\State Unlabeled dataset (\eg ESConv): $\sD_{U}$
\State Pseudo-label data constraint: $S$
\State Initial Classifier: $\mC_{\theta}$
\Ensure 
\State $i \gets 0$
\State $\sD^{i} \gets \sD_{L}^{tr}$
\State $\mC_{i} \gets $ fine-tuning $\mC_{\theta}$ on $\sD^{i}$ 
\While{$\mC_{i}$ does not have the best performance on $\sD_{L}^{va}$}
\State Predict labels on $\sD_{U}$ using $\mC_{i}$
\State Select prediction results by constraint $S$
\State Construct pseudo-labeled dataset $\sD^{i}_{PL}$ using selected data
\State $\sD^{i+1} \gets \sD^{i} \cup \sD^{i}_{PL}$
\State $\mC_{i+1} \gets$ fine-tuning $\mC_{i}$ on $\sD^{i+1}$ 
\State $i \gets i+1$
\EndWhile
\end{algorithmic}
\end{algorithm*}

\subsection{More Experiments Results}
\begin{table*}[h]
\centering
\resizebox{\textwidth}{!}{%
\begin{tabular}{r|cccccccccccc}
\hline
\multicolumn{1}{c|}{}                        & \multicolumn{4}{c|}{\textbf{DREAM}}                                                        & \multicolumn{4}{c|}{\textbf{ESConv}}                                                      & \multicolumn{4}{c}{\textbf{MSC}}                                     \\ \hline
\multicolumn{1}{c|}{\multirow{2}{*}{Metric}} & \multicolumn{2}{c}{Voting}        & IgnoreEqual     & \multicolumn{1}{c|}{Cont2Cat}        & \multicolumn{2}{c}{Voting}        & IgnoreEqual     & \multicolumn{1}{c|}{Cont2Cat}       & \multicolumn{2}{c}{Voting}        & IgnoreEqual     & Cont2Cat       \\
\multicolumn{1}{c|}{}                        & Pearson         & Spearman        & Point-Biserial  & \multicolumn{1}{c|}{IAA}             & Pearson         & Spearman        & Point-Biserial  & \multicolumn{1}{c|}{IAA}            & Pearson         & Spearman        & Point-Biserial  & IAA            \\ \hline
\multicolumn{1}{l|}{}                        & \multicolumn{12}{c}{\textbf{Relevance}}                                                                                                                                                                                                                       \\ \hline
\ourmethod              & \textbf{0.294*} & \textbf{0.334*} & \textbf{0.363}  & \multicolumn{1}{c|}{\textbf{0.369}}  & \textbf{0.312*} & \textbf{0.343*} & \textbf{0.402}  & \multicolumn{1}{c|}{\textbf{0.337}} & \textbf{0.257*} & \textbf{0.308*} & \textbf{0.316}  & \textbf{0.330} \\
-$p(c_i, c_j, r_t)$                          & 0.184           & 0.157           & 0.216           & \multicolumn{1}{c|}{0.312}           & 0.148           & 0.146           & 0.209           & \multicolumn{1}{c|}{0.284}          & 0.137           & 0.151           & 0.176           & 0.289          \\
-$p(c_i, r_t)$                               & 0.229           & 0.303*          & 0.335*          & \multicolumn{1}{c|}{0.347}           & 0.294*          & 0.328*          & 0.362           & \multicolumn{1}{c|}{0.327}          & 0.204           & 0.256*          & 0.292           & 0.318          \\
-self-training                               & 0.285*          & 0.325*          & 0.351*          & \multicolumn{1}{c|}{0.358}           & 0.302*          & 0.340*          & 0.387*          & \multicolumn{1}{c|}{0.336}          & 0.247*          & 0.299*          & 0.304*          & 0.324          \\
$\rightarrow$ MaxCI                               & 0.087          & 0.075          & 0.101          & \multicolumn{1}{c|}{0.302}           & 0.095          & 0.079          & 0.104          & \multicolumn{1}{c|}{0.277}          & 0.133          & 0.119          & 0.161          & 0.271          \\ 
$\rightarrow$ Preced2                               & 0.150          & 0.128          & 0.177          & \multicolumn{1}{c|}{0.303}           & 0.114          & 0.107          & 0.163          & \multicolumn{1}{c|}{0.280}          & 0.105          & 0.121          & 0.146          & 0.272          \\
\hline
\multicolumn{1}{l|}{}                        & \multicolumn{12}{c}{\textbf{Specificity}}                                                                                                                                                                                                                     \\ \hline
\ourmethod              & \textbf{0.328*} & \textbf{0.434*} & \textbf{0.390*} & \multicolumn{1}{c|}{\textbf{0.360*}} & \textbf{0.324*} & \textbf{0.379*} & \textbf{0.411*} & \multicolumn{1}{c|}{\textbf{0.359}} & \textbf{0.321*} & \textbf{0.356*} & \textbf{0.400*} & \textbf{0.355} \\
-$p(c_i, c_j, r_t)$                          & 0.162           & 0.244           & 0.190           & \multicolumn{1}{c|}{0.303}           & 0.116           & 0.140           & 0.166           & \multicolumn{1}{c|}{0.300}          & 0.193           & 0.176           & 0.229           & 0.310          \\
-$p(c_i, r_t)$                               & 0.307*          & 0.413*          & 0.346*          & \multicolumn{1}{c|}{0.347}           & 0.304*          & 0.351*          & 0.395*          & \multicolumn{1}{c|}{0.334}          & 0.302*          & 0.344*          & 0.387*          & 0.342          \\
-self-training                               & 0.325*          & 0.430*          & 0.384*          & \multicolumn{1}{c|}{0.351}           & 0.308*          & 0.360*          & 0.406*          & \multicolumn{1}{c|}{0.348}          & 0.317*          & 0.353*          & \textbf{0.400*} & 0.351          \\ 
$\rightarrow$ MaxCI                               & 0.085          & 0.083          & 0.102          & \multicolumn{1}{c|}{0.282}           & 0.091          & 0.144          & 0.132          & \multicolumn{1}{c|}{0.293}          & 0.140          & 0.156          & 0.175 & 0.296          \\ 
$\rightarrow$ Preced2                               & 0.052          & 0.072          & 0.142          & \multicolumn{1}{c|}{0.274}           & 0.103          & 0.121          & 0.158          & \multicolumn{1}{c|}{0.274}          & 0.135          & 0.142          & 0.213          & 0.304          \\
\hline
\multicolumn{1}{l|}{}                        & \multicolumn{12}{c}{\textbf{Empathy}}                                                                                                                                                                                                                         \\ \hline
\ourmethod              & \textbf{0.131}  & \textbf{0.252*} & \textbf{0.211}  & \multicolumn{1}{c|}{\textbf{0.325}}  & \textbf{0.186*} & \textbf{0.208*} & \textbf{0.302*} & \multicolumn{1}{c|}{\textbf{0.317}} & \textbf{0.131*} & \textbf{0.201*} & \textbf{0.292*} & \textbf{0.314} \\
$-p(c_i, c_j, r_t)$                          & 0.012           & 0.021           & 0.022           & \multicolumn{1}{c|}{0.273}           & 0.053           & 0.021           & 0.048           & \multicolumn{1}{c|}{0.254}          & 0.031           & 0.032           & 0.037           & 0.277          \\
$-p(c_i, r_t)$                               & 0.094           & 0.155           & 0.170           & \multicolumn{1}{c|}{0.296}           & 0.106           & 0.138           & 0.259           & \multicolumn{1}{c|}{0.287}          & 0.094           & 0.112           & 0.264           & 0.296          \\
-self-training                               & 0.113           & 0.231           & 0.200           & \multicolumn{1}{c|}{0.313}           & 0.151           & 0.172           & 0.281*          & \multicolumn{1}{c|}{0.307}          & 0.107           & 0.177           & 0.291*          & 0.304          \\ 
$\rightarrow$ MaxCI                               & -0.007          & -0.025          & -0.005          & \multicolumn{1}{c|}{0.251}           & 0.009          & -0.005          & 0.025          & \multicolumn{1}{c|}{0.254}          & 0.057          & 0.065          & 0.064 & 0.287          \\ 
$\rightarrow$ Preced2                               & 0.052          & 0.083          & 0.103          & \multicolumn{1}{c|}{0.263}           & 0.063          & 0.036          & 0.073          & \multicolumn{1}{c|}{0.259}          & 0.051          & 0.058          & 0.103          & 0.284          \\
\hline
\multicolumn{1}{l|}{}                        & \multicolumn{12}{c}{\textbf{Consistency}}                                                                                                                                                                                                                     \\ \hline
\ourmethod              & \textbf{0.206}  & \textbf{0.234*} & \textbf{0.222}  & \multicolumn{1}{c|}{\textbf{0.317}}  & \textbf{0.216}  & \textbf{0.238*} & \textbf{0.287}  & \multicolumn{1}{c|}{\textbf{0.337}} & \textbf{0.214}  & \textbf{0.231*} & \textbf{0.208}  & \textbf{0.315} \\
-$p(c_i, c_j, r_t)$                          & 0.056           & 0.030           & 0.085           & \multicolumn{1}{c|}{0.257}           & 0.113           & 0.118           & 0.143           & \multicolumn{1}{c|}{0.291}          & 0.131           & 0.180           & 0.144           & 0.295          \\
-$p(c_i, r_t)$                               & 0.193           & 0.201           & 0.208           & \multicolumn{1}{c|}{0.301}           & 0.202           & 0.227           & 0.278           & \multicolumn{1}{c|}{0.323}          & 0.170           & 0.200           & 0.201           & 0.309          \\
-self-training                               & 0.204           & 0.231*          & 0.215           & \multicolumn{1}{c|}{0.315}           & 0.210           & 0.232           & 0.283           & \multicolumn{1}{c|}{0.335}          & 0.189           & 0.215           & 0.205           & \textbf{0.315} \\ 
$\rightarrow$ MaxCI                               & -0.023          & 0.023          & -0.031          & \multicolumn{1}{c|}{0.265}           & 0.077          & 0.045          & 0.052          & \multicolumn{1}{c|}{0.282}          & 0.090          & 0.046          & 0.104 & 0.247          \\ 
$\rightarrow$ Preced2                               & 0.073          & 0.052          & 0.097          & \multicolumn{1}{c|}{0.271}           & 0.092          & 0.115          & 0.133          & \multicolumn{1}{c|}{0.287}          & 0.145          & 0.173          & 0.156          & 0.294          \\
\hline
\multicolumn{1}{l|}{}                        & \multicolumn{12}{c}{\textbf{Overall}}                                                                                                                                                                                                                         \\ \hline
\ourmethod              & \textbf{0.331*} & \textbf{0.422*} & \textbf{0.511*} & \multicolumn{1}{c|}{\textbf{0.595}}  & \textbf{0.287*} & \textbf{0.339*} & \textbf{0.411*} & \multicolumn{1}{c|}{\textbf{0.568}} & \textbf{0.331*} & \textbf{0.401*} & \textbf{0.492*} & \textbf{0.569} \\
-$p(c_i, c_j, r_t)$                          & 0.192           & 0.231           & 0.303           & \multicolumn{1}{c|}{0.517}           & 0.115           & 0.121           & 0.161           & \multicolumn{1}{c|}{0.483}          & 0.179           & 0.235           & 0.272           & 0.526          \\
-$p(c_i, r_t)$                               & 0.303*          & 0.396*          & 0.496*          & \multicolumn{1}{c|}{0.571}           & 0.262*          & 0.314*          & 0.403*          & \multicolumn{1}{c|}{0.548}          & 0.316*          & 0.380*          & 0.473*          & 0.546          \\
-self-training                               & 0.326*          & 0.414*          & 0.503*          & \multicolumn{1}{c|}{0.586}           & 0.284*          & 0.331*          & 0.407*          & \multicolumn{1}{c|}{\textbf{0.568}} & 0.324*          & 0.387*          & 0.488*          & 0.562 \\
$\rightarrow$ MaxCI                               & 0.203          &   0.147         & 0.250          & \multicolumn{1}{c|}{0.490}           & 0.048          & 0.087          & 0.058          & \multicolumn{1}{c|}{0.473}          & 0.086          & 0.116          & 0.112 & 0.480                \\ 
$\rightarrow$ Preced2                               & 0.172          & 0.158          & 0.183          & \multicolumn{1}{c|}{0.358}           & 0.103          & 0.095          & 0.135          & \multicolumn{1}{c|}{0.263}          & 0.126          & 0.131          & 0.157          & 0.301          \\
\hline
\end{tabular}%
}
\caption{Ablation results on three datasets.}
\label{tab:ablation_study_corr}
\end{table*}
\begin{table*}[!h]
\centering
\resizebox{\textwidth}{!}{%
\begin{tabular}{ll|cccc|cccc|cccc}
\hline
\multicolumn{2}{c|}{\textbf{}}       & \multicolumn{4}{c|}{\textbf{DREAM}}                                 & \multicolumn{4}{c|}{\textbf{ESConv}}                                & \multicolumn{4}{c}{\textbf{MSC}}                                    \\ \hline
\multicolumn{2}{c|}{\multirow{2}{*}{Metric}}  & \multicolumn{2}{c}{Voting}        & IgnoreEqual    & Cont2Cat       & \multicolumn{2}{c}{Voting}        & IgnoreEqual    & Cont2Cat       & \multicolumn{2}{c}{Voting}        & IgnoreEqual    & Cont2Cat       \\
\multicolumn{2}{c|}{}                         & Pearson         & Spearman        & Point-Biserial & IAA            & Pearson         & Spearman        & Point-Biserial & IAA            & Pearson         & Spearman        & Point-Biserial & IAA            \\ \hline
\multicolumn{14}{c}{\textbf{Relevance}}  \\ \hline
\multirow{5}{*}{Reference-based} & BLEU       & 0.021           & 0.018           & 0.027          & 0.246          & -0.047          & -0.065          & -0.053         & 0.216          & 0.076           & 0.065           & 0.074          & 0.222          \\
                                 & ROUGE      & -0.005          & -0.008          & -0.015         & 0.272          & 0.039           & 0.020           & 0.045          & 0.243          & 0.091           & 0.087           & 0.090          & 0.249          \\
                                 & METEOR     & 0.028           & 0.033           & 0.043          & 0.262          & 0.097           & 0.097           & 0.081          & 0.243          & 0.013           & 0.036           & 0.087          & 0.241          \\
                                 & BERTScore  & -0.004          & -0.010          & -0.003         & 0.260          & 0.085           & 0.069           & 0.092          & 0.246          & 0.021           & 0.002           & 0.031          & 0.239          \\
                                 & BLEURT     & -0.022          & -0.032          & -0.030         & 0.257          & 0.025           & 0.022           & 0.026          & 0.246          & 0.074           & 0.089           & 0.094          & 0.249          \\ \hline
\multirow{5}{*}{Reference-free} & PPL      & 0.033           & 0.097          & 0.043          & 0.292          & -0.040           & -0.031           & -0.073          & 0.246          & -0.046          & -0.047           & -0.053          & 0.245          \\ 
                                & GRADE      & 0.004           & -0.005          & 0.035          & 0.248          & 0.013           & 0.021           & 0.030          & 0.248          & -0.003          & 0.012           & 0.049          & 0.243          \\
                                 & DEAM       & -0.090          & -0.053          & -0.121         & 0.273          & -0.011          & 0.039           & -0.011         & 0.257          & -0.012          & -0.032          & -0.007         & 0.253          \\
                                 & DEnsity       &  0.117         &  0.112         &  0.149        &  0.286         &  0.080         &   0.099          & 0.095         & 0.268          &  0.030         & 0.030          &  0.026        &  0.258         \\
                                 & ChatGPT    & 0.036           & 0.024           & 0.088          & 0.284          & -0.002          & -0.018          & 0.096          & 0.250          & 0.083           & 0.084           & 0.109          & 0.271          \\
                                 & GPT4       & 0.049           & 0.038           & 0.083          & 0.263          & -0.002          & -0.023          & 0.097          & 0.251          & 0.039           & 0.083           & 0.110          & 0.277          \\
                                 & \ourmethod & \textbf{0.294*} & \textbf{0.334*} & \textbf{0.363} & \textbf{0.369} & \textbf{0.312*} & \textbf{0.343*} & \textbf{0.402} & \textbf{0.337} & \textbf{0.257*} & \textbf{0.308*} & \textbf{0.316} & \textbf{0.330} \\ \hline \hline
\multicolumn{14}{c}{\textbf{Specificity}}  \\ \hline
\multirow{5}{*}{Reference-based} & BLEU       & -0.079          & -0.066          & 0.023           & 0.220          & 0.051           & 0.037           & 0.063           & 0.239          & 0.041           & 0.016           & 0.004           & 0.239          \\
                                 & ROUGE      & 0.043           & 0.052           & 0.049           & 0.259          & 0.046           & 0.043           & 0.061           & 0.253          & -0.001          & -0.019          & 0.001           & 0.251          \\
                                 & METEOR     & 0.066           & 0.061           & 0.081           & 0.251          & 0.093           & 0.040           & 0.080           & 0.259          & 0.002           & 0.011           & 0.001           & 0.263          \\
                                 & BERTScore  & 0.043           & 0.065           & 0.044           & 0.249          & 0.067           & 0.072           & 0.095           & 0.265          & 0.036           & 0.033           & 0.046           & 0.266          \\
                                 & BLEURT     & 0.041           & 0.038           & 0.050           & 0.249          & -0.003          & -0.004          & 0.003           & 0.258          & 0.031           & 0.020           & 0.046           & 0.261          \\ \hline
\multirow{5}{*}{Reference-free} & PPL      & 0.057           & 0.047           & 0.066           & 0.263          & -0.038           & 0.002           & -0.062           & 0.266          & -0.001           & -0.068           & -0.032           & 0.249          \\
                                & GRADE      & 0.047           & 0.031           & 0.058           & 0.251          & 0.057           & 0.047           & 0.071           & 0.271          & 0.063           & 0.041           & 0.069           & 0.268          \\
                                 & DEAM       & 0.063           & 0.029           & 0.078           & 0.255          & 0.039           & 0.015           & 0.059           & 0.274          & 0.048           & 0.039           & 0.047           & 0.253          \\
                                 & DEnsity       &  0.060         &  0.066         &  0.066        &  0.265         &  0.124         &   0.118          & 0.165         & 0.296          &  0.112         & 0.082          &  0.125        &  0.287         \\
                                 & ChatGPT    & 0.026           & 0.017           & 0.033           & 0.266          & 0.042           & 0.039           & 0.054           & 0.276          & 0.070           & 0.082           & 0.085           & 0.286          \\
                                 & GPT4       & 0.057           & 0.038           & 0.057           & 0.260          & 0.017           & 0.014           & 0.016           & 0.258          & 0.069           & 0.068           & 0.080           & 0.263          \\
                                 & \ourmethod & \textbf{0.328*} & \textbf{0.434*} & \textbf{0.390*} & \textbf{0.360} & \textbf{0.324*} & \textbf{0.379*} & \textbf{0.411*} & \textbf{0.359} & \textbf{0.321*} & \textbf{0.356*} & \textbf{0.400*} & \textbf{0.355} \\ \hline \hline
\multicolumn{14}{c}{\textbf{Consistency}}  \\ \hline
\multirow{5}{*}{Reference-based} & BLEU       & 0.045          & 0.050           & 0.041          & 0.225          & 0.049          & 0.009           & 0.014          & 0.241          & 0.047          & -0.007         & 0.015          & 0.216          \\
                                 & ROUGE      & 0.002          & -0.014          & -0.002         & 0.242          & -0.023         & -0.041          & -0.028         & 0.247          & 0.002          & -0.008         & -0.006         & 0.229          \\
                                 & METEOR     & -0.022         & -0.016          & -0.027         & 0.231          & 0.015          & 0.011           & 0.018          & 0.254          & -0.021         & -0.027         & -0.044         & 0.231          \\
                                 & BERTScore  & -0.001         & 0.012           & -0.003         & 0.228          & -0.036         & -0.033          & -0.042         & 0.258          & -0.005         & -0.015         & -0.009         & 0.229          \\
                                 & BLEURT     & 0.008          & 0.007           & 0.009          & 0.244          & -0.031         & -0.017          & -0.038         & 0.252          & -0.006         & -0.017         & -0.010         & 0.232          \\ \hline
\multirow{5}{*}{Reference-free} & PPL      & 0.014          &  0.025           & 0.012          & 0.260          & 0.017          & 0.011           & 0.020          & 0.270          & 0.012          & 0.027          & 0.053          & 0.275          \\ 
                                & GRADE      & 0.041          & 0.014           & 0.042          & 0.262          & 0.057          & 0.008           & 0.071          & 0.268          & 0.048          & 0.027          & 0.079          & 0.267          \\
                                 & DEAM       & 0.035          & 0.019           & 0.039          & 0.259          & 0.061          & 0.002           & 0.082          & 0.269          & 0.043          & 0.019          & 0.066          & 0.237          \\
                                 & DEnsity       &  -0.015        &  -0.025        &  -0.014       &  0.240         &  0.043         &   0.046          & 0.058         & 0.282          &  0.076         & 0.044          &  0.103        &  0.259         \\
                                 & ChatGPT    & 0.067          & 0.034           & 0.041          & 0.261          & 0.083          & 0.055           & 0.082          & 0.273          & 0.099          & 0.101          & 0.119          & 0.278          \\
                                 & GPT4       & 0.073          & 0.043           & 0.047          & 0.264          & 0.088          & 0.048           & 0.065          & 0.271          & 0.103          & 0.124          & 0.132          & 0.273          \\
                                 & \ourmethod & \textbf{0.206} & \textbf{0.234*} & \textbf{0.222} & \textbf{0.317} & \textbf{0.216} & \textbf{0.238*} & \textbf{0.287} & \textbf{0.337} & \textbf{0.214} & \textbf{0.231*} & \textbf{0.208} & \textbf{0.315} \\ \hline \hline
\multicolumn{14}{c}{\textbf{Empathy}}  \\ \hline
\multirow{5}{*}{Reference-based} & BLEU       & 0.019           & 0.058           & 0.012           & 0.225          & 0.069           & 0.050           & 0.024           & 0.205          & 0.019           & -0.020          & 0.007           & 0.203          \\
                                 & ROUGE      & 0.031           & 0.028           & 0.041           & 0.233          & 0.050           & 0.041           & 0.044           & 0.215          & 0.021           & 0.018           & 0.009           & 0.215          \\
                                 & METEOR     & 0.043           & 0.031           & 0.054           & 0.234          & 0.041           & 0.029           & 0.052           & 0.235          & 0.016           & 0.015           & 0.007           & 0.214          \\
                                 & BERTScore  & 0.065           & 0.077           & 0.103           & 0.238          & -0.028          & 0.004           & -0.035          & 0.238          & 0.032           & 0.042           & 0.052           & 0.220          \\
                                 & BLEURT     & 0.011           & 0.005           & 0.011           & 0.231          & 0.112           & 0.117           & 0.161           & 0.239          & 0.076           & 0.077           & 0.111           & 0.235          \\ \hline
\multirow{5}{*}{Reference-free} & PPL      & 0.018           & -0.031           & 0.016           & 0.242          & 0.021           & -0.018           & 0.008           & 0.245          & 0.030           & 0.032           & 0.061           & 0.248         \\ 
                                & GRADE      & 0.063           & 0.048           & 0.024           & 0.247          & 0.046           & 0.030           & 0.075           & 0.243          & 0.079           & 0.063           & 0.094           & 0.247          \\
                                 & DEAM       & 0.111           & 0.073           & 0.168           & 0.247          & 0.053           & 0.036           & 0.086           & 0.247          & 0.041           & 0.021           & 0.074           & 0.242          \\
                                 & DEnsity       &  0.039        &  0.013        &  0.056       &  0.259         &  -0.034        &  -0.036          & -0.038         & 0.253          &  0.055         &  0.047          &  0.059        &  0.286         \\
                                 & ChatGPT    & 0.110           & 0.084           & 0.106           & 0.251          & 0.062           & 0.045           & 0.074           & 0.243          & 0.109           & 0.084           & 0.112           & 0.241          \\
                                 & GPT4       & 0.105           & 0.085           & 0.132           & 0.246          & 0.029           & 0.026           & 0.041           & 0.228          & 0.105           & 0.082           & 0.127           & 0.238          \\
                                 & \ourmethod & \textbf{0.131} & \textbf{0.252*} & \textbf{0.211} & \textbf{0.325} & \textbf{0.186} & \textbf{0.208*} & \textbf{0.302*} & \textbf{0.317} & \textbf{0.131} & \textbf{0.201} & \textbf{0.292*} & \textbf{0.314} \\ \hline \hline
\multicolumn{14}{c}{\textbf{Overall}}  \\ \hline
\multirow{5}{*}{Reference-based} & BLEU       & 0.019           & 0.058           & 0.011           & 0.444          & 0.069           & 0.050           & 0.005           & 0.434          & 0.019           & -0.019          & 0.007           & 0.422          \\
                                 & ROUGE      & -0.031          & -0.028          & -0.040          & 0.453          & -0.030          & -0.041          & -0.044          & 0.445          & -0.011          & -0.018          & -0.010          & 0.435          \\
                                 & METEOR     & -0.043          & -0.031          & -0.053          & 0.454          & 0.041           & 0.029           & 0.052           & 0.455          & 0.006           & 0.015           & 0.007           & 0.435          \\
                                 & BERTScore  & 0.065           & 0.077           & 0.103           & 0.458          & -0.028          & 0.004           & -0.035          & 0.458          & 0.032           & 0.042           & 0.052           & 0.440          \\
                                 & BLEURT     & 0.011           & 0.005           & 0.011           & 0.451          & -0.112          & -0.117          & -0.161          & 0.439          & 0.076           & 0.077           & 0.112           & 0.445          \\ \hline
\multirow{5}{*}{Reference-free} & PPL      & 0.034           & 0.010           & 0.032           & 0.454          & 0.045           & 0.105           & 0.100           & 0.480          & 0.023           & -0.038           & -0.022           & 0.436          \\ 
                                & GRADE      & 0.054           & 0.033           & 0.012           & 0.454          & 0.023           & 0.011           & 0.004           & 0.436          & 0.088           & 0.050           & 0.105           & 0.442          \\
                                 & DEAM       & 0.111           & 0.107           & 0.168           & 0.467          & 0.013           & 0.010           & 0.005           & 0.442          & 0.042           & 0.021           & 0.074           & 0.442          \\
                                 & DEnsity       &  0.011         &  0.009         &  0.023        &  0.462         &  0.038         &    0.100          & 0.064         & 0.483          &  0.076         & 0.045          &  0.091        &  0.465         \\
                                 & ChatGPT    & 0.153           & 0.101           & 0.113           & 0.460          & 0.052           & 0.055           & 0.041           & 0.463          & 0.129           & 0.125           & 0.181           & 0.481          \\
                                 & GPT4       & 0.159           & 0.157           & 0.141           & 0.486          & 0.048           & 0.062           & 0.042           & 0.471          & 0.131           & 0.103           & 0.119           & 0.486          \\
                                 & \ourmethod & \textbf{0.331*} & \textbf{0.422*} & \textbf{0.511*} & \textbf{0.595} & \textbf{0.287*} & \textbf{0.339*} & \textbf{0.411*} & \textbf{0.568} & \textbf{0.331*} & \textbf{0.401*} & \textbf{0.492*} & \textbf{0.569} \\ \hline
\end{tabular}%
}
\caption{Correlations on all dimensions between automatic evaluation metrics and human judgements on three different datasets (DREAM, ESConv, MSC). Inter-annotator agreement (IAA) is computed using Krippendorff's alpha. PPL represents perplexity.}
\label{tab:metric_corr_all_dims}
\end{table*} 

\paragraph{Evaluation on All Dimensions.} Besides Relevance and Overall dimensions, we also present correlations on Specificity, Empathy, and Consistency in Table~\ref{tab:ablation_study_corr} and ~\ref{tab:metric_corr_all_dims}. Additionally, \ourmethod shows higher performance on specificity and overall than relevance. Specificity measures the degree to which responses are generated to the dialogue history. The high specificity often correlates with elevated relevance, as specific responses are typically more relevant. Regarding the overall dimension, it is posited that annotators tend to favor responses having high relevance and specificity, perceiving them as indicative of superior overall quality. In terms of consistency and empathy dimensions, \ourmethod also surpasses baseline metrics, although with less distinction compared to its achievements in relevance, specificity, and overall assessment.

\begin{figure*}
    \centering
    \includegraphics[width=0.7\textwidth]{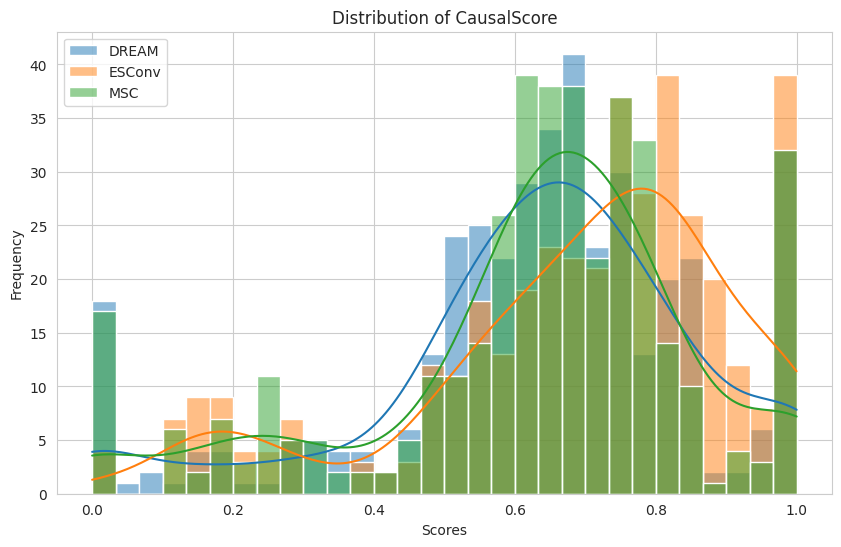}
    \caption{Distribution of \ourmethod on three datasets with a kernel density estimate to smooth the distribution. }
    \label{fig:causal_score_dist}
\end{figure*}

\paragraph{Distribution of \ourmethod.} \ourmethod is bounded between 0 and 1, where a higher score indicates greater relevance between the dialogue history and the response. A score of 0 indicates complete irrelevance, implying no causal connection between the response and the preceding dialogue. Conversely, a score of 1 signifies the highest relevance, demonstrating a direct and significant causal link. As depicted in Figure~\ref{fig:causal_score_dist}, the distribution of \ourmethod across different datasets covers the full spectrum of scores from 0 to 1. 

\paragraph{Out-of-Domain Evaluation.} 

\begin{table*}[h]
\centering
\resizebox{\textwidth}{!}{%
\begin{tabular}{lcccccccccccc}
\hline
\multicolumn{1}{l|}{}                                  & \multicolumn{4}{c|}{\textbf{DREAM}}                                                                     & \multicolumn{4}{c|}{\textbf{ESConv}}                                                                    & \multicolumn{4}{c}{\textbf{MSC}}                                                   \\ \hline
\multicolumn{1}{l|}{\multirow{2}{*}{\textbf{Metrics}}} & \multicolumn{2}{c}{\textbf{Voting}}  & \textbf{IgnoreEqual}    & \multicolumn{1}{c|}{\textbf{Cont2Cat}} & \multicolumn{2}{c}{\textbf{Voting}}  & \textbf{IgnoreEqual}    & \multicolumn{1}{c|}{\textbf{Cont2Cat}} & \multicolumn{2}{c}{\textbf{Voting}}  & \textbf{IgnoreEqual}    & \textbf{Cont2Cat} \\
\multicolumn{1}{l|}{}                                  & \textbf{Pearson} & \textbf{Spearman} & \textbf{Point-Biserial} & \multicolumn{1}{c|}{\textbf{IAA}}      & \textbf{Pearson} & \textbf{Spearman} & \textbf{Point-Biserial} & \multicolumn{1}{c|}{\textbf{IAA}}      & \textbf{Pearson} & \textbf{Spearman} & \textbf{Point-Biserial} & \textbf{IAA}      \\ \hline
\multicolumn{13}{c}{\textbf{Relevance}}                                                                                                                                                                                                                                                                                                                         \\ \hline
\multicolumn{1}{l|}{PPL}                               & 0.033            & 0.097             & 0.043                   & \multicolumn{1}{c|}{0.292}             & -0.040           & -0.031            & -0.073                  & \multicolumn{1}{c|}{0.246}             & -0.046           & -0.047            & -0.053                  & 0.245             \\
\multicolumn{1}{l|}{GRADE}                             & 0.004            & -0.005            & 0.035                   & \multicolumn{1}{c|}{0.248}             & 0.013            & 0.021             & 0.030                   & \multicolumn{1}{c|}{0.248}             & -0.003           & 0.012             & 0.049                   & 0.243             \\
\multicolumn{1}{l|}{DEAM}                              & -0.090           & -0.053            & -0.121                  & \multicolumn{1}{c|}{0.273}             & -0.011           & 0.039             & -0.011                  & \multicolumn{1}{c|}{0.257}             & -0.012           & -0.032            & -0.007                  & 0.253             \\
\multicolumn{1}{l|}{DEnsity}                           & 0.117            & 0.112             & 0.149                   & \multicolumn{1}{c|}{0.286}             & 0.080            & 0.099             & 0.095                   & \multicolumn{1}{c|}{0.268}             & 0.030            & 0.030             & 0.026                   & 0.258             \\
\multicolumn{1}{l|}{ChatGPT}                           & 0.036            & 0.024             & 0.088                   & \multicolumn{1}{c|}{0.284}             & -0.002           & -0.018            & 0.096                   & \multicolumn{1}{c|}{0.250}             & 0.083            & 0.084             & 0.109                   & 0.271             \\
\multicolumn{1}{l|}{GPT4}                              & 0.049            & 0.038             & 0.083                   & \multicolumn{1}{c|}{0.263}             & -0.002           & -0.023            & 0.097                   & \multicolumn{1}{c|}{0.251}             & 0.039            & 0.083             & 0.110                   & 0.277             \\
\multicolumn{1}{l|}{\ourmethod-DREAM}                  & \textbf{0.294*}  & \textbf{0.334*}   & \textbf{0.363}          & \multicolumn{1}{c|}{\textbf{0.369}}    & 0.101            & 0.103             & 0.195                   & \multicolumn{1}{c|}{0.285}             & 0.107            & 0.132             & 0.176                   & 0.295             \\
\multicolumn{1}{l|}{\ourmethod-ESConv}                 & 0.094            & 0.101             & 0.135                   & \multicolumn{1}{c|}{0.285}             & \textbf{0.312*}  & \textbf{0.343*}   & \textbf{0.402}          & \multicolumn{1}{c|}{\textbf{0.337}}    & 0.108            & 0.109             & 0.163                   & 0.285             \\
\multicolumn{1}{l|}{\ourmethod-MSC}                    & 0.109            & 0.124             & 0.154                   & \multicolumn{1}{c|}{0.294}             & 0.145            & 0.122             & 0.232                   & \multicolumn{1}{c|}{0.304}             & \textbf{0.257*}  & \textbf{0.308*}   & \textbf{0.316}          & \textbf{0.330}    \\
\multicolumn{1}{l|}{\ourmethod}                        & \textbf{0.294*}  & \textbf{0.334*}   & \textbf{0.363}          & \multicolumn{1}{c|}{\textbf{0.369}}    & \textbf{0.312*}  & \textbf{0.343*}   & \textbf{0.402}          & \multicolumn{1}{c|}{\textbf{0.337}}    & \textbf{0.257*}  & \textbf{0.308*}   & \textbf{0.316}          & \textbf{0.330}    \\ \hline
\end{tabular}%
}
\caption{Out-of-Domain Performance of \ourmethod.}
\label{tab:out_of_domain_performance}
\end{table*}
As discussed in the Limitations Section~\ref{sec:limitations}, the efficacy of \ourmethod is limited by the availability of human-annotated cause-effect relationships for the training of conditional independence classifiers.  In the absence of such annotations, there is a potential for diminished performance when \ourmethod is applied to unseen domains. Table~\ref{tab:out_of_domain_performance} provides a quantitative evaluation of \ourmethod's out-of-domain performance. For instance, \ourmethod-ESConv, which is trained on the CGDIALOG+(ESConv) subset, has a diminished performance on the MSC and DREAM datasets. \ourmethod-DREAM and \ourmethod-MSC have similar observations. Although there is a drop in performance within the Out-of-Domain setting, \ourmethod maintains equivalent or superior results relative to baseline models.

\subsection{Qualitative Study}
\label{apx:qualitative_study}
In this section we present more evaluation examples in Table~\ref{tab:case_study_apx_1}, ~\ref{tab:case_study_apx_2}, ~\ref{tab:case_study_apx_3}, ~\ref{tab:case_study_apx_4} to provide a more intuitive assessment of \ourmethod. In Table~\ref{tab:case_study_apx_5}, we demonstrate that our method can assign a score nearing zero to the completely irrelevant responses (\ie generated by \blenderbot), and assign a score nearing one for relevant responses (provided by humans).

\begin{table*}[h]
\centering
\resizebox{\textwidth}{!}{%
\begin{tabular}{llccccc}
\hline
W:           & \multicolumn{6}{l}{How was the game, Bill? Did you enjoy it?}                                                                     \\
M:           & \multicolumn{6}{l}{No, it was not interesting at all.}                                                                            \\
W:           & \multicolumn{6}{l}{That's too bad. Football games are usually exciting.}                                                          \\
M:           & \multicolumn{6}{l}{Not last night. Some of the players didn't know what they were doing. In fact, one of them was just terrible.} \\
W:           & \multicolumn{6}{l}{Well, which team was the winner?}                                                                              \\
M:           & \multicolumn{6}{l}{The Tigers, they won the game 3-1.}                                                                            \\ \hline
             & \multicolumn{1}{l|}{}                                                         & \multicolumn{5}{c}{Evaluation on Relevance}       \\
Responses:   & \multicolumn{1}{l|}{}                                                         & Human  & \ourmethod  & GPT4  & DEAM   & BERTScore \\ \hline
Human:       & \multicolumn{1}{l|}{Were you happy about the score?}                          & 9      & 0.643       & 5     & 0.997  & 1.0       \\
\alpaca:     & \multicolumn{1}{l|}{That's great! It's always exciting when the Tigers win.}  & 7      & 0.624       & 2     & 0.997  & 0.866     \\
\blenderbot: & \multicolumn{1}{l|}{The Tigers are one of my favorite baseball teams.}        & 3      & 0.576       & 4     & 0.996  & 0.854     \\
ConSTrain:   & \multicolumn{1}{l|}{It was a close game, but the Tigers won.}                 & 4      & 0.622       & 4     & 0.996  & 0.871     \\ \hline
\end{tabular}%
}
\caption{One example with evaluation results on human judgement, \ourmethod, GPT4, DEAM, and BERTScore. We use voting schema on all pairwise comparisons to get human scores.}
\label{tab:case_study_apx_1}
\end{table*}
\begin{table*}[h]
\centering
\resizebox{\textwidth}{!}{%
\begin{tabular}{llccccc}
\hline
supporter:   & \multicolumn{6}{l}{Hello there, how are you feeling at the moment?}                                                                                                                                                                                                                                   \\
seeker:      & \multicolumn{6}{l}{I am feeling okay, but I have been better}                                                                                                                                                                                                                                         \\
supporter:   & \multicolumn{6}{l}{Is there anything specific that you'd like to talk about with me?}                                                                                                                                                                                                                 \\
seeker:      & \multicolumn{6}{l}{I am having a lot of conflict with my boss at work.}                                                                                                                                                                                                                               \\
supporter:   & \multicolumn{6}{l}{That is not good. Are you thinking of looking for new work?}                                                                                                                                                                                                                       \\
seeker:      & \multicolumn{6}{l}{Honestly maybe. It is exhausting}                                                                                                                                                                                                                                                  \\
supporter:   & \multicolumn{6}{l}{I had a similar situation myself a few years ago and I moved on. I've never regretted it.}                                                                                                                                                                                         \\
seeker:      & \multicolumn{6}{l}{It is just a little scary since I don't have anything else set up}                                                                                                                                                                                                                 \\
supporter:   & \multicolumn{6}{l}{Is your boss's behaviour unreasonable enough to justify making a complaint about him to HR?}                                                                                                                                                                                       \\
seeker:      & \multicolumn{6}{l}{I have and they just contacted her boss. Nothing changed after.}                                                                                                                                                                                                                   \\ \hline
             & \multicolumn{1}{l|}{}                                                                                                                                                                                                                                 & \multicolumn{5}{c}{Evaluation on Relevance}   \\
Responses:   & \multicolumn{1}{l|}{}                                                                                                                                                                                                                                 & Human & \ourmethod & GPT4 & DEAM  & BERTScore \\ \hline
Human:       & \multicolumn{1}{l|}{\begin{tabular}[c]{@{}l@{}}It might be worth following up with. \\ There are organisations that could give you free advice.\end{tabular}}                                                                                         & 7     & 0.647      & 4    & 0.995 & 1.0       \\ \hline
\alpaca:     & \multicolumn{1}{l|}{\begin{tabular}[c]{@{}l@{}}It sounds like you've been through a lot with your boss. \\ Have you thought about what kind of job you would look for \\ if you decided to move on? Is there anything I can do to help?\end{tabular}} & 9     & 0.964      & 4    & 0.997 & 0.851     \\ \hline
\blenderbot: & \multicolumn{1}{l|}{That's good to hear. I hope it all works out for you.}                                                                                                                                                                            & 2     & 0.457      & 2    & 0.996 & 0.857     \\ \hline
ConSTrain:   & \multicolumn{1}{l|}{Contacting her boss is a step in the right direction.}                                                                                                                                                                            & 7     & 0.939      & 3    & 0.990 & 0.858     \\ \hline
\end{tabular}%
}
\caption{One example with evaluation results.}
\label{tab:case_study_apx_2}
\end{table*}
\begin{table*}[h]
\centering
\resizebox{\textwidth}{!}{%
\begin{tabular}{llccccc}
\hline
supporter:   & \multicolumn{6}{l}{Hi. How are you doing?}                                                                                                                                                                                                                                                                                  \\
seeker:      & \multicolumn{6}{l}{I am very sad}                                                                                                                                                                                                                                                                                           \\
supporter:   & \multicolumn{6}{l}{oh, what happened? why are you sad?}                                                                                                                                                                                                                                                                     \\
seeker:      & \multicolumn{6}{l}{I think I am just feeling down because of the pandemic}                                                                                                                                                                                                                                                  \\
supporter:   & \multicolumn{6}{l}{\begin{tabular}[c]{@{}l@{}}ok, i understand. this pandemic has been really tough. \\ have you tried to take your mind off of the pandemic by spending time on your hobbies?\end{tabular}}                                                                                                                \\
seeker:      & \multicolumn{6}{l}{I have but they aren't the same without my friends}                                                                                                                                                                                                                                                      \\
supporter:   & \multicolumn{6}{l}{\begin{tabular}[c]{@{}l@{}}that's true. times like these really show us the value of our friends. \\ have you talked to your friends recently? either in person or through phone/video calls?\end{tabular}}                                                                                              \\
seeker:      & \multicolumn{6}{l}{We have done a few video calls but it just makes me more sad that I can't see them in person}                                                                                                                                                                                                            \\
supporter:   & \multicolumn{6}{l}{\begin{tabular}[c]{@{}l@{}}okay. at lease seeing them in the video should have brought you some relief. \\ Are you planning to get vaccinated when you are able to? And what about your friends? once you are vaccinated, \\ it will help you meet your friends in person.\end{tabular}}                 \\
seeker:      & \multicolumn{6}{l}{Yes we are planning to get it as soon as possible but that is hard in our state}                                                                                                                                                                                                                         \\
supporter:   & \multicolumn{6}{l}{oh okay. Have you tried checking with your friends to see if you could meet in person but still maintain social distance and wear masks, etc.}                                                                                                                                                           \\
seeker:      & \multicolumn{6}{l}{None of them are really comfortable with that sadly.}                                                                                                                                                                                                                                                    \\ \hline
             & \multicolumn{1}{l|}{}                                                                                                                                                                                                                                                       & \multicolumn{5}{c}{Evaluation on Relevance}   \\
Responses:   & \multicolumn{1}{l|}{}                                                                                                                                                                                                                                                       & Human & \ourmethod & GPT4 & DEAM  & BERTScore \\ \hline
Human:       & \multicolumn{1}{l|}{\begin{tabular}[c]{@{}l@{}}oh okay, have you discussed anything with your friends to see \\ if you find any alternative way of meeting?\end{tabular}}                                                                                                   & 9     & 0.788      & 5    & 0.997 & 1.0       \\ \hline
\alpaca:     & \multicolumn{1}{l|}{\begin{tabular}[c]{@{}l@{}}That's understandable. Have you thought about reaching out to other people \\ in the same situation and trying to organize a group vaccination? \\ That might make it easier for all of you to get vaccinated.\end{tabular}} & 9     & 0.803      & 5    & 0.995 & 0.876     \\ \hline
\blenderbot: & \multicolumn{1}{l|}{I don't understand why they don't all of them are comfortable with it.}                                                                                                                                                                                 & 6     & 0.676      & 2    & 0.996 & 0.840     \\ \hline
ConSTrain:   & \multicolumn{1}{l|}{What do you do for a job? Do you have the money?}                                                                                                                                                                                                       & 1     & 0.501      & 1    & 0.994 & 0.862     \\ \hline
\end{tabular}%
}
\caption{One example with evaluation results.}
\label{tab:case_study_apx_3}
\end{table*}
\begin{table*}[h]
\centering
\resizebox{\textwidth}{!}{%
\begin{tabular}{llccccc}
\hline
             & \multicolumn{6}{l}{…}                                                                                                                                                                                                                                                                                \\
Speaker 2:   & \multicolumn{6}{l}{\begin{tabular}[c]{@{}l@{}}it hasn't happened yet! we're still packing and getting ready but i'm already exhausted. \\ how long did it take you to recover from your move?\end{tabular}}                                                                                          \\
Speaker 1:   & \multicolumn{6}{l}{It took about 3 weeks to fully recover from all of the packing and unpacking. It's not fun at all!}                                                                                                                                                                               \\
Speaker 2:   & \multicolumn{6}{l}{\begin{tabular}[c]{@{}l@{}}yes and because it's like, your HOME, there's just no peaceful place to recover from the chaos!\\  what had prompted your most recent move?\end{tabular}}                                                                                              \\
Speaker 1:   & \multicolumn{6}{l}{\begin{tabular}[c]{@{}l@{}}Exactly! And I was ready for a bigger house, I needed more space. \\ I bet you're going to need even more time recovering since you had a long mountain hike!\end{tabular}}                                                                            \\
Speaker 2:   & \multicolumn{6}{l}{\begin{tabular}[c]{@{}l@{}}I totally hear you on space. nobody likes to be cramped. and yeah, \\ i'm definitely sore from the hike but it was good to clear my head before the big move! \\ what do you do to relax or unwind when you're in a stressful situation?\end{tabular}} \\
Speaker 1:   & \multicolumn{6}{l}{\begin{tabular}[c]{@{}l@{}}I love to just take a stroll around the mall on my lunch break. The mall is usually empty so it's kind of peaceful. \\ Other than yellow do you have anything else in mind for colors?\end{tabular}}                                                   \\
Speaker 2:   & \multicolumn{6}{l}{\begin{tabular}[c]{@{}l@{}}that sounds really nice. fresh air! and yes, I want to paint the living room a really vibrant purple. \\ my couch is orange and I think it would look cool!\end{tabular}}                                                                              \\
Speaker 1:   & \multicolumn{6}{l}{Yes that would look really good together! I can't wait to get a new couch, the dogs have messed up the couch I have now.}                                                                                                                                                         \\
Speaker 2:   & \multicolumn{6}{l}{aw man thats such a bummer. what color couch would you get do you think?}                                                                                                                                                                                                         \\
Speaker 1:   & \multicolumn{6}{l}{I think I want a dark grey couch, something where stains from the dogs doesn't show so easily. I love them, but they are messy!}                                                                                                                                                  \\
Speaker 2:   & \multicolumn{6}{l}{\begin{tabular}[c]{@{}l@{}}big dogs often are! dark colors hide stains better so thats smart. \\ I have cats so I had to get a special fabric couch so their claws couldnt ruin it.\end{tabular}}                                                                                 \\
Speaker 1:   & \multicolumn{6}{l}{I am going to send you a list of my best moving tips that will save you time and trouble.}                                                                                                                                                                                        \\ \hline
             & \multicolumn{1}{l|}{}                                                                                                                                                            & \multicolumn{5}{c}{Evaluation on Relevance}                                                                       \\
Responses:   & \multicolumn{1}{l|}{}                                                                                                                                                            & Human                & \ourmethod               & GPT4              & DEAM               & BERTScore              \\ \hline
Human:       & \multicolumn{1}{l|}{\begin{tabular}[c]{@{}l@{}}Oh, that would be excellent! Thank you. \\ I am not excited to move next week as it is so much work.\end{tabular}}                & 10                   & 0.695                    & 5                 & 0.991              & 1.0                    \\ \hline
\alpaca:     & \multicolumn{1}{l|}{Hey there, what are you up to? I'm going to work at the mall.}                                                                                               & 6                    & 0.626                    & 5                 & 0.995              & 0.859                  \\ \hline
\blenderbot: & \multicolumn{1}{l|}{That sounds like a great idea! Thank you so much!}                                                                                                           & 8                    & 0.530                    & 1                 & 0.993              & 0.882                  \\ \hline
ConSTrain:   & \multicolumn{1}{l|}{That would be great. Moving can be so stressful and time-consuming.}                                                                                         & 10                   & 0.730                    & 5                 & 0.997              & 0.895                  \\ \hline
\end{tabular}%
}
\caption{One example with evaluation results.}
\label{tab:case_study_apx_4}
\end{table*}

\begin{table*}[h]
\centering
\resizebox{\textwidth}{!}{%
\begin{tabular}{llccccc}
\hline
             & \multicolumn{6}{l}{……}                                                                                                                   \\
W:           & \multicolumn{6}{l}{Have we agreed where we're going to meet John on Sunday?}                                                             \\
M:           & \multicolumn{6}{l}{Well, I talked to the others and nobody seemed to have any idea where they wanted to go.}                             \\
W:           & \multicolumn{6}{l}{I thought we were going to that restaurant.., what was it?}                                                           \\
M:           & \multicolumn{6}{l}{... the McDonald's at the end of Beijing Road--I suggested it a long time ago.}                                       \\
W:           & \multicolumn{6}{l}{Well--I think we should agree on that.}                                                                               \\
M:           & \multicolumn{6}{l}{OK, shall I phone the others now?}                                                                                    \\ \hline
             & \multicolumn{1}{l|}{}                                                                      & \multicolumn{5}{c}{Evaluation on Relevance} \\
Responses:   & \multicolumn{1}{l|}{}                                                                      & Human  & \ourmethod  & GPT4  & DEAM  & BLEU \\ \hline
Human:       & \multicolumn{1}{l|}{No, thanks. It is better for me to do that!}                           & 9      & 0.999       & 3     & 0.995 & 1.0  \\
\alpaca:     & \multicolumn{1}{l|}{Have we agreed where we're going to meet John on Sunday?}              & 3      & 0.475       & 1     & 0.001 & 0.0  \\
\blenderbot: & \multicolumn{1}{l|}{I don't know what that is, but I'm sure you can find something to do.} & 1      & 0.004       & 1     & 0.995 & 0.0  \\
ConSTrain:   & \multicolumn{1}{l|}{Yes, I'd like to. Are you free this afternoon?}                        & 3      & 0.554       & 2     & 0.995 & 0.0  \\ \hline
\end{tabular}%
}
\caption{One example with evaluation results. \ourmethod can output a score close to zero for the irrelevant response generated by \blenderbot.}
\label{tab:case_study_apx_5}
\end{table*}

\end{document}